\documentclass{article}

\usepackage[preprint]{corl_2026} 

\usepackage[utf8]{inputenc}   
\usepackage[T1]{fontenc}      
 
\usepackage[table,dvipsnames]{xcolor}  
\usepackage{graphicx}
\usepackage{hyperref}          
\usepackage{url}               
 
\usepackage{amsmath}
\usepackage{amsfonts}          
\usepackage{nicefrac}          
 
\usepackage{array}
\usepackage{booktabs}          
\usepackage{tabularx}
\usepackage{multirow}
\usepackage{makecell}
\usepackage{wrapfig}
\usepackage{bbm}
 
\usepackage{algorithm}
\usepackage{algpseudocode}
 
\usepackage{relsize}
\usepackage{xspace}
\usepackage{pifont}            
 
\usepackage{natbib}
\usepackage{multibib}
\usepackage{titletoc}
 
\usepackage{duckuments}
 
\definecolor{mgray}{gray}{.9}
\definecolor{dgray}{gray}{.5}
 
\newcolumntype{P}[1]{>{\centering\arraybackslash}p{#1}}
 
\title{Scaling Self-Play for End-to-End Driving}

\author{
  \textbf{Luke Rowe}$^{1,2}$, \textbf{Roger Girgis}$^{1,3,4}$, \textbf{Rodrigue de Schaetzen}$^{1,2,4}$, \\
  \textbf{Daphne Cornelisse}$^{5}$, \textbf{Alaap Grandhi}$^{1,6}$, \textbf{Felix Heide}$^{4,7}$, \textbf{Eugene Vinitsky}$^{5}$, \\
  \textbf{Christopher Pal}$^{1,2,3}$, \textbf{Liam Paull}$^{1,2}$ \\
  $^1$Mila, $^2$Université de Montréal, $^3$Polytechnique Montréal, $^4$Torc Robotics, \\
  $^5$NYU Tandon School of Engineering, $^6$McMaster University, $^7$Princeton University 
  \\ {\normalsize \textbf{\url{https://montrealrobotics.ca/gigapixel}}}
  \vspace{-5mm}
}

\begin{document}
\maketitle


\begin{abstract}
  End-to-end autonomous driving models are typically trained on offline human-demonstration datasets that provide limited state coverage and often no closed-loop feedback, making them prone to compounding errors when deployed in closed-loop and brittle to long-tail agent interactions. To overcome these limitations, we propose an alternative strategy for training end-to-end driving models: large-scale self-play directly from pixels in simulation. While prior self-play approaches have shown promising transfer to real-world driving, they typically assume vectorized Bird’s-Eye-View (BEV) observations that are incompatible with end-to-end policies operating directly on sensor observations. To this end, we introduce \emph{Gigapixel}, a high-throughput batched driving simulator with perspective rendering, enabling scalable self-play directly from pixel observations. Rather than targeting compute-costly photorealistic sensor simulation, Gigapixel renders a simplified bounding-box world that preserves essential scene structure while achieving throughput at 50k agent steps per second. Since direct pixel-space self-play RL is prohibitively sample-inefficient at end-to-end model scale, we propose \emph{self-play DAgger} training: we train pixel-based policies in self-play via on-policy distillation from a privileged RL teacher. To bridge the sim-to-real gap, we subsequently transfer the self-play trained policies to real-world sensor data through lightweight perception adaptation. Policies trained in Gigapixel and adapted to real-world sensor data achieve competitive performance on the HUGSIM and NAVSIM-v2 benchmarks without human trajectory supervision. Moreover, scaling self-play training yields proportional gains in policy performance, establishing self-play as a practical and scalable strategy for training end-to-end models.
\end{abstract}


\section{Introduction}

Autonomous driving has undergone a decisive shift towards end-to-end models that directly map sensor inputs to planning outputs \cite{chib2023recent, chen2024end, chitta2022transfuser, hu2023planning, jiang2023vad, hwang2024emma, rowe2025poutine}. Beyond architectural simplicity, end-to-end formulations are inherently scalable: they optimize the planning objective directly, and performance improves predictably with increasing data and model capacity \cite{zheng2024data, naumann2025data, baniodeh2025scaling}. Realizing this scaling potential, however, requires a principled and scalable training approach. Behavior cloning (BC) of human driving logs remains the dominant training paradigm~\cite{le2022survey, zheng2024data, naumann2025data, baniodeh2025scaling, rowe2025poutine}, but it suffers from structural limitations. Logged datasets provide limited state coverage, leaving policies brittle once they reach states outside nominal human driving~\cite{ljungbergh2024neuroncap}. Moreover, the absence of closed-loop interaction during training induces covariate shift, leading to compounding errors at deployment~\cite{karkus2025beyond}. These shortcomings are intrinsic to BC and do not disappear merely from collecting more data~\cite{ross2011reduction, codevilla2019exploring}.


Self-play in simulation offers a principled alternative. Agents learn through \textit{closed-loop} interaction with copies of themselves, generating a diverse experience curriculum during training. Closed-loop interaction enables learning the consequences of its own actions, which improves robustness to compounding errors at test time~\cite{karkus2025beyond}. Furthermore, unlike offline datasets, which are fixed and costly to expand, self-play state coverage scales directly with compute, enabling arbitrarily large and targeted experience generation. Realizing this benefit at scale requires a high-throughput simulator to cheaply generate on-policy experience. Recent batched simulators such as Gigaflow~\cite{cusumano2025robust} and PufferDrive~\cite{pufferdrive2025github} demonstrate throughputs exceeding hundreds of thousands of steps per second (SPS), enabling robust closed-loop learning from self-play~\cite{cornelissehuman, cornelisse2025building, zhang2024asymmetric, cusumano2025robust, pufferdrive2025github}. 

\begin{wrapfigure}{r}{0.5\textwidth}
    \centering
    \vspace{-18pt}
    \includegraphics[width=0.48\textwidth]{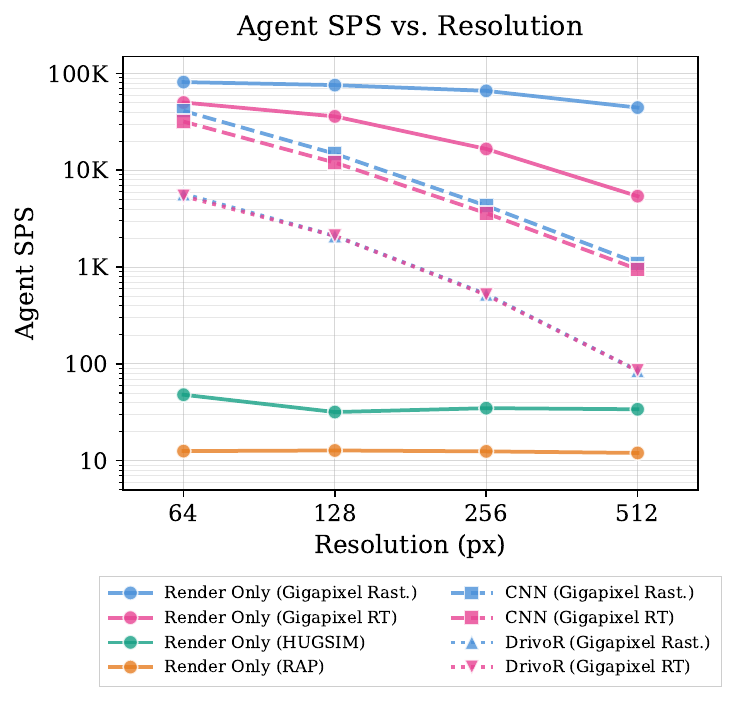}
    \vspace{-3mm}
    \caption{\textbf{Gigapixel Throughput vs.\ Resolution.}
    Agent steps per second (SPS) across rendering resolutions and policy architectures on 1 NVIDIA A100L GPU.
    \textit{Render Only} isolates renderer throughput without policy forward or backward passes.
    \textit{CNN} is a simple CNN, and \textit{DrivoR} \cite{kirby2026driving} is a transformer-based architecture. CNN and DrivoR throughputs are reported in an RL training loop.
    The gap between the rasterizer (\textit{Rast.}) and ray-tracer (\textit{RT}) throughput narrows as model complexity increases,
    indicating that Gigapixel rendering ceases to be the throughput bottleneck. For comparison, we report HUGSIM and RAP throughputs using their public code.}
    \label{fig:training_sps}
    \vspace{-12pt}
\end{wrapfigure}

However, these simulators produce vectorized BEV observations incompatible with end-to-end policies that must act from raw sensor inputs. To overcome this limitation, we introduce \textit{Gigapixel}, a high-throughput batched driving simulator that extends PufferDrive with ray-traced and rasterized perspective rendering~\cite{rosenzweig2024high}, enabling \textbf{scalable self-play training directly from pixel observations at 50k agent SPS on 1 GPU}, with throughput scaling near-linearly in the number of GPUs. Rather than simulating photorealistic sensors, Gigapixel renders perspective views of a simplified bounding-box world that preserves essential scene geometry and interaction fidelity.

While Gigapixel enables high-throughput batched simulation from pixels, direct self-play RL at end-to-end model scale remains prohibitively sample-inefficient due to the cost of policy forward and backward passes (Figure~\ref{fig:training_sps}). As a more sample-efficient alternative, we introduce \textit{self-play DAgger}: a privileged teacher policy is first trained via RL on vectorized observations, then distilled into a pixel-based student via DAgger~\cite{ross2011reduction, agarwal2024policy} in self-play. This preserves the benefits of closed-loop self-play while dramatically reducing sample complexity. Self-play DAgger additionally enables training a trajectory-output policy, matching the output format of standard end-to-end planners \cite{hu2023planning, jiang2023vad, wang2025alpamayo, liao2025diffusiondrive}, rather than the control outputs of typical RL-based planners \cite{jaeger2025carl, cusumano2025robust}. Then, to deploy these policies on real sensor data, we isolate the sim-to-real gap to perception: the planning head already captures the closed-loop behaviors learned in self-play, so we adapt only the perception module to map real sensor inputs into the planning head's latent representation.

We evaluate Gigapixel self-play trained policies in simulation and on real-world benchmarks. Our approach achieves state-of-the-art performance on the closed-loop HUGSIM benchmark~\cite{zhou2025hugsim} and competitive performance on the pseudo-closed-loop NAVSIM-v2 benchmark~\cite{cao2025pseudo}, \textit{without any human trajectory supervision}. Empirically, self-play training scales consistently: increasing training experience yields proportional gains in policy performance. We summarize our contributions: \textbf{1.} We propose self-play as a scalable alternative to offline behavior cloning for training end-to-end driving models. We enable this through self-play DAgger, a sample-efficient alternative to direct self-play RL from pixels. \textbf{2.} We introduce Gigapixel, a high-throughput batched perspective-rendering simulator that enables pixel-based self-play at scale. \textbf{3.} We show that the robust driving behavior learned via self-play in simulation transfers to robust driving from real-world observations through lightweight sim-to-real perception adaptation.

Together, these contributions establish self-play as a scalable and practical strategy for training end-to-end driving models—one that addresses the structural limitations of offline behavior cloning and opens a path to continual improvement through synthetic experience.
\section{Related work}
\label{sec:related_work}

\paragraph{End-to-End Autonomous Driving.} The growing availability of large-scale datasets of human driving logs \cite{caesar2020nuscenes, caesar2021nuplan, sun2020scalability, xu2025wod, arai2025covla, ghilotti2026truckdrive} have enabled tremendous progress in behavior cloning for end-to-end driving \cite{le2022survey}, spanning transformer-based \cite{hu2023planning, jiang2023vad, chitta2022transfuser, weng2024drive, li2024hydra, kirby2026driving}, diffusion-based \cite{liao2025diffusiondrive, xing2025goalflow}, and vision-language model-based \citep{hwang2024emma, renz2025simlingo, zhou2026autovla, rowe2025poutine} methods. Despite this progress, behavior cloning suffers from structural limitations that induce brittle behavior when deployed in closed-loop \cite{karkus2025beyond, codevilla2019exploring, ross2011reduction, de2019causal, dauner2023pdmclosed}. To address these limitations, prior works have explored direct closed-loop training in simulators via RL either from scratch \cite{kiran2021deep, chen2019model, toromanoff2020end, chekroun2023gri, zhang2022rethinking} or as a post-training stage \cite{gao2026rad, ni2025recondreamer}. Other works propose DAgger-based methods \cite{ross2011reduction, zhang2025closed} with privileged state-based experts \cite{zhang2016query, prakash2020exploring, garcia2025road},  test-time adaptation \cite{zhao2026bridgesim}, or rollouts in a latent world model \cite{popov2024mitigating}. However, these prior methods predominantly focus on the single-agent paradigm and therefore train on environments with limited behavioral diversity. Moreover, prior works typically train in slow simulators such as CARLA \cite{dosovitskiy2017carla, coelho2023rlad, toromanoff2020end, osinski2020simulation} or neural reconstruction-based simulators \cite{zhou2025hugsim, ni2025recondreamer} that limit scalability. Our method is a closed-loop DAgger-based method, but unlike prior works, we propose a \textit{multi-agent} training paradigm that leverages the benefits of scalable self-play in a fast simulator \cite{cusumano2025robust} to learn a robust end-to-end driving policy.

\paragraph{Autonomous Driving Simulators.} The space of driving simulators is characterized by a tension between throughput and representational fidelity. At one extreme, abstract simulators operate on vectorized BEV state representations \cite{caesar2021nuplan, gulino2024waymax, pufferdrive2025github, vinitsky2022nocturne, gpudrive2025}, sacrificing perceptual realism for speed. At the other end, photorealistic simulators render high-fidelity sensor data to minimize the sim-to-real gap, either via handcrafted assets \cite{wymann2000torcs, dosovitskiy2017carla, li2022metadrive}, neural reconstruction methods \cite{yang2023unisim, tonderski2024neurad, zhou2025hugsim, alpasim_2025}, or generative models \cite{hu2023gaia, russell2025gaia, yang2025resim, yang2025drivearena, yan2025drivingsphere}. However, this comes at a dramatic cost of training throughput (steps per second or SPS). We argue that Gigapixel occupies a productive middle ground: it retains the throughput necessary to experiment with self-play and closed-loop training at scale while providing perspective-view observations that make it suitable for training end-to-end driving policies. The representational approach that most closely mirrors Gigapixel is RAP \cite{feng2025rap}, which also renders simplified bounding box worlds. However, as shown in Figure~\ref{fig:training_sps}, RAP rendering is significantly slower than Gigapixel due to the lack of batched GPU rendering support \cite{shacklett2023extensible, rosenzweig2024high}, and RAP utilizes these simplified renderings for data augmentation rather than closed-loop training.

\paragraph{Self-Play for Driving.} Self-play RL algorithms are data-hungry, requiring billions of steps of experience to reach (super) human-level performance. The appeal of self-play lies in the nature of the experience it generates. Unlike behavior cloning, which learns from a fixed and narrow distribution of human demonstrations, self-play agents are paired with themselves and are given an objective to optimize. The distribution of encountered behaviors continually shifts as they learn, surfacing many safety-critical interactions that are vanishingly rare in human driving logs \cite{liu2024curse, zhang2024asymmetric, cusumano2025robust}. The effectiveness of this paradigm has its roots in competitive and cooperative games~\cite{silver2017alphazero,go2017,population2018,diplomacy2023}. Thus far, the results in driving have been confined to policies operating on vectorized observations \cite{cornelissehuman, cornelisse2025building, chang2025spacer, cusumano2025robust, wang2026nomad, distelzweig2026beyond, qiu2026heterogeneous}. We go beyond prior works by enabling large-scale self-play directly from pixel observations. Moreover, we propose a closed-loop self-play training procedure based on DAgger rather than RL to enable sample-efficient self-play learning at end-to-end model scale.

\section{Self-Play for End-to-End Driving}

\begin{figure}[t]
\label{fig:system}
    \centering
    \includegraphics[width=\linewidth]{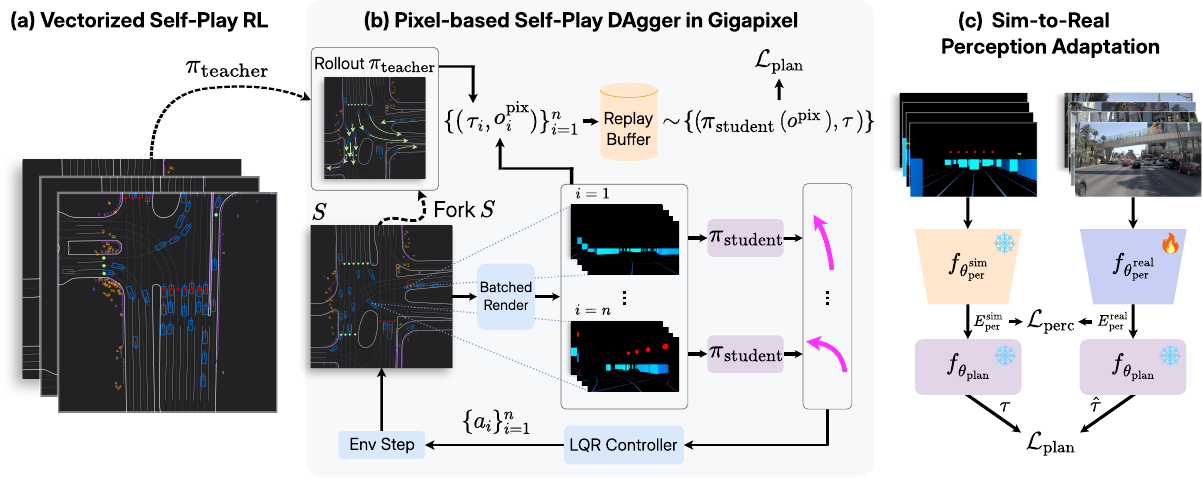}
    \vspace{-18 pt}
    \caption{\textbf{Self-Play for End-to-End Driving.} We propose self-play training for end-to-end driving in three stages: \textbf{(a)}~a vectorized teacher is trained with self-play RL in an abstract simulator; \textbf{(b)}~a pixel-based student is distilled via self-play DAgger in Gigapixel, with all agents student-controlled and trajectory targets generated by parallel teacher rollouts (Sec.~\ref{subsec:self-play-dagger}); and \textbf{(c)}~the student is adapted to real images by finetuning only the perception backbone on paired sim-real observations (Sec.~\ref{subsec:sim2real}).}
    \label{fig:gigapixel_system_fig}
\vspace{-10 pt}
\end{figure}


\paragraph{Problem formulation.} Our goal is to learn a robust end-to-end driving policy $\pi(\tau | I, C)$ that outputs a trajectory $\tau$ from raw image observations $I$ and additional ego context $C$ (\textit{e.g.}, ego state and navigation command). We represent a trajectory as a sequence of waypoints $\tau = \{(x_t, y_t, \theta_t)\}_{t=1}^{H}$, where $(x_t, y_t)$ denotes the ego-centric position and $\theta_t$ the heading of the ego vehicle at future timestep $t$, over a finite planning horizon $H$. A low-level controller maps $\tau$ to a sequence of control actions, of which the first action $a$ is applied before replanning at the next timestep in a closed-loop (receding-horizon) fashion. We assume $\pi$ decomposes into a perception backbone $f_{\theta_{\text{per}}}$ that produces perception features $E_{\text{per}} = f_{\theta_{\text{per}}}(I, C)$, and a planning head $f_{\theta_{\text{plan}}}$ that maps these features to a trajectory, $\tau = f_{\theta_{\text{plan}}}(E_{\text{per}}, C)$. This decomposition is general and satisfied by a wide range of end-to-end architectures \cite{chitta2022transfuser, kirby2026driving, hu2023planning, jiang2023vad, wang2025alpamayo, liao2025diffusiondrive, renz2025simlingo}. We place no constraints on $f_{\theta_{\text{plan}}}$ beyond its input-output interface; it may be scoring-based \cite{li2024hydra, kirby2026driving}, diffusion-based \cite{liao2025diffusiondrive} , or regression-based \cite{chitta2022transfuser, renz2025simlingo}.


\paragraph{Online RL and DAgger.} Self-play is a \textit{closed-loop} training paradigm: the agent rolls out actions in the simulator, allowing it to learn the consequences of its own actions during training \cite{karkus2025beyond}. Two standard paradigms enable closed-loop learning in simulation. Online RL \cite{sutton1998reinforcement, sutton1999policy, mnih2015human, schulman2015trust, schulman2017proximal} trains a policy $\pi$ to maximize the expected discounted return $\mathbb{E}_{\pi}\!\left[\sum_t \gamma^t r_t\right]$ with discount factor $\gamma$, where $r_t$ is the reward obtained at step $t$. The policy is rolled out in the simulator and trained on the generated experience. While effective, online RL is sample-inefficient \cite{sutton1998reinforcement}, often requiring billions of environment steps to converge. This becomes prohibitive when the policy is a large end-to-end model, as each step incurs costly forward and backward passes through the model. Dataset Aggregation (DAgger) \cite{ross2011reduction} offers a more sample-efficient alternative when an expert policy is available. At each iteration, the student policy $\pi_{\text{student}}$ is rolled out in the environment, the visited states are labeled by an expert $\pi_{\text{expert}}$, and the student is trained to match the expert's actions on the student's own induced state distribution. Crucially, the student and expert may each act through its own observation function, denoted by $O_{\text{student}}$ and $O_{\text{expert}}$. 
The DAgger objective is:
\begin{equation}
\label{eq:single-agent-dagger}
\min_{\theta}\; \mathbb{E}_{S \sim d_{\pi_{\text{student}}}}\!\left[\mathcal{L}\bigl(\pi^{\theta}_{\text{student}}(O_{\text{student}}(S)),\, \pi_{\text{expert}}(O_{\text{expert}}(S))\bigr)\right],
\end{equation}
with action loss $\mathcal{L}$. Like RL, DAgger is closed-loop: the expectation is taken under $\pi_{\text{student}}$'s own state distribution $d_{\pi_{\text{student}}}$, so the student learns to recover from the states it actually visits. 

\subsection{Gigapixel}

A key obstacle to leveraging self-play for end-to-end driving is that existing simulators either natively expose \textit{vectorized} observations incompatible with end-to-end policies \cite{cusumano2025robust, vinitsky2022nocturne, pufferdrive2025github, caesar2021nuplan, gpudrive2025}, or are too slow to support self-play at scale \cite{dosovitskiy2017carla, zhou2025hugsim, yang2025drivearena, caesar2021nuplan}. Our central observation is that learning robust closed-loop end-to-end driving behavior does not require a photorealistic sensor simulator. Instead, it suffices to learn in a \textit{high-throughput} abstract simulator that preserves the essential information required for planning while natively supporting pixel observations. To this end, we introduce \textit{Gigapixel}, a driving simulator that enables scalable self-play directly from pixel observations. In Gigapixel, the global simulator state takes the form $S_t = (A_t, M, L_t)$, comprising agent bounding boxes $A_t$, static map polylines $M$, and traffic light states $L_t$. Prior abstract simulators implement vectorized observations $O_{\text{vec}}(S_t)$ \cite{pufferdrive2025github, cusumano2025robust}, which are cheap to produce but incompatible with end-to-end policies that consume raw sensor inputs. Gigapixel additionally implements $O_{\text{pixel}}(S_t)$, rendering $S_t$ into an ego-centric perspective view through a batched, GPU-accelerated renderer. Concretely, we extend the PufferDrive abstract simulator \cite{pufferdrive2025github} with the Madrona rendering engine \cite{shacklett2023extensible, rosenzweig2024high}, supporting both rasterization \cite{foley1996computer, feng2025rap} and ray tracing \cite{glassner1989introduction}. To sustain high throughput, scene elements are represented as simple primitives: agents (\textit{e.g.,} vehicles, pedestrians, cyclists) and static obstacles as cuboids, lane polylines as thin planar strips, and traffic lights as small spheres. Figure~\ref{fig:gigapixel_system_fig} shows ray-traced renderings; rasterized counterparts appear in the Appendix. Further details about the Gigapixel simulator---including the renderer, dynamics, action space, and reward functions---are deferred to the Appendix.



\subsection{Self-play DAgger}
\label{subsec:self-play-dagger}

Self-play RL from pixels faces two obstacles in our setting. First, training compute-intensive end-to-end architectures in Gigapixel shifts the per-step cost from rendering to policy forward and backward passes; combined with RL's sample inefficiency, this makes self-play RL from pixels prohibitively expensive. Second, end-to-end policies typically output trajectories $\tau$, whereas RL-trained driving policies typically output low-level controls  \cite{saxena2020driving, harmel2023scaling, jaeger2025carl, cusumano2025robust}. To address both, we propose \textit{self-play DAgger} training: pixel-based end-to-end policies are trained in self-play via on-policy distillation from a privileged RL teacher. Critically, RL is tractable for the vectorized teacher but prohibitive for the pixel-based student (Fig.~\ref{fig:training_sps}): the teacher is a lightweight policy over low-dimensional vectorized observations, making its forward/backward passes cheap, and self-play RL is known to scale to robust, naturalistic driving in high-throughput simulators~\cite{cusumano2025robust, cornelisse2025building}. We therefore confine RL to the teacher and distill its closed-loop behavior into the expensive pixel-based student. Distillation is far more sample-efficient than online RL (see Fig.~\ref{fig:sample-efficiency}), and rolling out the teacher in a forked parallel environment produces trajectory targets $\tau$ directly, sidestepping the waypoint-vs-controls mismatch.

\paragraph{Self-play DAgger vs. vanilla DAgger.}
Standard DAgger assumes a single learner whose visited states are labeled by an expert. In autonomous driving, this typically means the learner controls the ego vehicle while other agents follow log-replay or handcrafted behaviors. In our setting, every agent in the scene is controlled by the student policy $\pi_{\text{student}}$, so the induced state distribution is a function of all agents' joint behavior. The student is trained on states drawn from the self-play rollout distribution $S \sim d^{\,\text{self-play}}_{\pi_{\text{student}}}$, the marginal state distribution induced when $\pi_{\text{student}}$ controls all agents. This gives the self-play DAgger objective:
\[
\min_{\theta}\; \mathbb{E}_{S \sim d^{\,\text{self-play}}_{\pi_{\text{student}}}}\;\mathbb{E}_{i}\!\left[\mathcal{L}\bigl(\pi_{\text{student}}^{\theta}(O_{\text{student}}(S, i)),\, \pi_{\text{teacher}}(O_{\text{teacher}}(S, i))\bigr)\right],
\]
which is Eq.~\ref{eq:single-agent-dagger} with the self-play induced state distribution $d^{\,\text{self-play}}_{\pi_{\text{student}}}$ in place of the single-learner distribution and an added expectation over agents $i$. This yields two benefits over vanilla DAgger. First, because all agents are controlled by $\pi_{\text{student}}$, the scenarios encountered during training co-evolve with the policy, continually surfacing interesting multi-agent interactions that are rare in human driving logs. Second, as made explicit by the inner expectation over $i$, every agent in a joint rollout contributes training data, multiplying the training experience extracted per simulator step over vanilla DAgger. We now describe our specific instantiation of self-play DAgger.

\paragraph{Vectorized teacher training.} Self-play DAgger requires a teacher $\pi_{\text{teacher}}$ that is robust across the state distribution visited by the student and can be queried cheaply for supervision. We obtain our teacher by adapting Gigaflow~\cite{cusumano2025robust} which generates robust and naturalistic driving policies from self-play RL, albeit over vectorized BEV representations. We train a compact decentralized policy $\pi_{\text{teacher}}(a_{i} | o^{\text{vec}}_{i})$ on ego-centric vectorized observations $o^{\text{vec}}_{i} = O_{\text{vec}}(S, i)$. Training uses decentralized PPO~\cite{schulman2017proximal} with a linearly weighted multi-objective reward $R_i = \sum_{j=1}^{N_r} c_i^j R^j$ consisting of $N_r$ individual reward terms (such as collision, offroad, comfort, \textit{etc}.). On each episode reset, the coefficients $\mathbf{c}_i := (c_i^1, \dots, c_i^{N_r})$ are randomized per agent $i$ and provided to the policy as conditioning~\cite{rowectrlsim}, so a single policy models different driving \textit{personas} (e.g., cautious vs. aggressive). This induces a diverse population of agents during self-play training, improving the policy's robustness. 

\paragraph{Pixel-based student training.} Given a trained teacher $\pi_{\text{teacher}}$, we distill it into the pixel-based student $\pi_{\text{student}}$ via self-play DAgger, where $\pi_{\text{teacher}}$ provides \textit{trajectory-level} supervision. Every agent $i$ is controlled by $\pi_{\text{student}}$, which acts on the ego-centric pixel observation $o^{\text{pix}}_{i} = O_{\text{pix}}(S, i)$ and outputs a trajectory tracked by an LQR controller; the resulting joint behavior induces the global simulator state $S$. To generate supervision at $S$, we fork a parallel simulator instance initialized at $S$ and roll out the teacher in self-play for $H$ steps, acting on $o^{\text{vec}}_{i}$. This yields a per-agent trajectory target $\tau_i$ for every agent $i$ in the scene, giving us the self-play DAgger objective:
\[
\min_{\theta}\; \mathbb{E}_{S \sim d^{\,\text{self-play}}_{\pi_{\text{student}}}}\;\mathbb{E}_{i}\!\left[\mathcal{L}_{\text{plan}}\bigl(\pi_{\text{student}}^{\theta}(o^{\text{pix}}_{i}),\, \tau_i\bigr)\right],
\]
where $\mathcal{L}_{\text{plan}}$ scores the student's predicted trajectory $\hat{\tau}_i = \pi_{\text{student}}^{\theta}(o^{\text{pix}}_{i})$ against the target $\tau_i$.\footnote{We leave $\mathcal{L}_{\text{plan}}$ deliberately general: it may be any planning loss compatible with the planning head $f_{\theta_{\text{plan}}}$.} Because the teacher is conditioned on a per-agent reward-preference vector $\mathbf{c}_i$, the target $\tau_i$ is persona-specific. We therefore condition the student on the same $\mathbf{c}_i$ and match it to the teacher during distillation, so the target is consistent with the student's inputs; sampling $\mathbf{c}_i$ additionally exposes the student to diverse behaviors during training. We suppress $\mathbf{c}_i$ in the notation above for brevity; full conditioning details are provided in the Appendix.

\begin{table}[t]
    \centering
    \small
    \setlength{\tabcolsep}{1.5pt}
    \begin{tabular}{@{}l|c|ccccc|ccccc@{}}
    \toprule
    & Human & \multicolumn{5}{c}{RC} & \multicolumn{5}{c}{HD-Score} \\
    Method & Trajectory? & E & M & H & X & Avg. & E & M & H & X & Avg.\\
    \midrule
    LTF~\cite{chitta2022transfuser} & \ding{51}
    &
    67.8 & 35.1 & 26.2 & 40.5 & 38.9 &
    58.9 & 18.0 & \phantom{0}9.8 & 25.9 & 23.7 \\
    \midrule
    VAD~\cite{jiang2023vad} & \ding{51}
    &
    51.3 & 31.1 & 25.3 & 26.5 & 31.4 &
    36.3 & \phantom{0}9.5 & \phantom{0}8.0 & 11.5 & 13.4 \\
    \midrule
    UniAD~\cite{hu2023planning} & \ding{51}
    &
    78.4 & 60.5 & 33.6 & 17.8 & 45.9 &
    64.9 & 45.8 & 20.6 & \phantom{0}6.6 & 32.7 \\
    \midrule
    \rowcolor[HTML]{EEE5F4}
    \multicolumn{12}{c}{\textit{Regression-based (Single-Mode) DrivoR Planner}} \\
    \cmidrule{1-12}
     DrivoR-Reg~\cite{kirby2026driving} & \ding{51}
    &
    75.8 & 28.0 & 32.3 & 46.8 & 40.5 &
    60.3 & 8.0 & 11.5 & 25.8 & 20.7 \\
    \cmidrule{1-12}
    \rowcolor[rgb]{0.9,0.95,1}
    Gigapixel-DrivoR-Reg  & \ding{55}
    &
    84.1 & 54.3 & 29.8 & 41.1 & 49.2 &
    65.8 & 38.8 & 13.1 & 24.6 & 33.2 \\
    \midrule
    \rowcolor[HTML]{EEE5F4}
    \multicolumn{12}{c}{\textit{Scoring-based (Multimodal) DrivoR Planner}} \\
    \cmidrule{1-12}
    DrivoR~\cite{kirby2026driving} & \ding{51}
    & 80.9 & 50.5 & 33.8 & 47.1 & \underline{49.8} 
    & 73.3 & 34.6 & 18.8 & 32.5 & 35.7 \\
    \cmidrule{1-12}
    DrivoR (w/ SimScale)~\cite{kirby2026driving} & \ding{51}
    & 70.4 & 54.4 & 28.2 & 39.2 & 46.4 
    & 64.8 & 50.3 & 16.8 & 25.8 & \underline{38.1} \\
    \cmidrule{1-12}
    \rowcolor[rgb]{0.9,0.95,1}
    Gigapixel-DrivoR  & \ding{55}
    & 78.6 & 60.7 & 32.7 & 35.4 & \textbf{50.1} 
    & 67.4 & 51.9 & 19.1 & 21.6 & \textbf{38.5} \\
    \bottomrule
    \end{tabular}
    \vspace{3mm}
    \caption{\textbf{Closed-loop evaluation in the photorealistic HUGSIM simulator \citep{zhou2025hugsim}}. Scores are reported per difficulty level (E=Easy, M=Medium, H=Hard, X=Extreme). RC=Route Completion, and HD-Score is the HUGSIM Driving Score. \textit{Human Trajectory?} denotes whether the human trajectory is used as a supervision target during training. Best method is \textbf{bolded}; second-best \underline{underlined}.}
    \vspace{-5mm}
    \label{tab:hugsim}
\end{table}

\subsection{Sim-to-real Perception Adaptation}
\label{subsec:sim2real}
Self-play DAgger yields a student $\pi_{\text{student}}$ that drives robustly in closed loop on Gigapixel's abstract renderings $o^{\text{pix}}_{i}$, but deploying on real sensor data requires closing the gap between these renderings and real camera images. We frame this as an image-to-image \textit{perceptual adaptation} task \cite{isola2017image, bousmalis2018using}: rather than retraining the full policy, we adapt only the perception stack so that real images map into the same latent representation the planning head already acts on. We curate a dataset of paired observations $\mathcal{D}_{\text{paired}} = \{(o^{\text{real}}, o^{\text{pix}})\}$, where $o^{\text{pix}}$ is the Gigapixel rendering of the abstract state reconstructed from the real log corresponding to $o^{\text{real}}$. From the trained student weights $\theta = (\theta_{\text{per}}, \theta_{\text{plan}})$ we instantiate two copies sharing the frozen planning head $f_{\theta_{\text{plan}}}$: a frozen teacher $\pi^{\text{sim}}$ on $o^{\text{pix}}$ and a trainable student $\pi^{\text{real}}$ on $o^{\text{real}}$, whose perception backbone $f_{\theta_{\text{per}}^{\text{real}}}$ is the only module updated. 
$\pi^{\text{sim}}$ supervises $\pi^{\text{real}}$ at both the output and feature levels via the planning loss $\mathcal{L}_{\text{plan}}$ used in self-play DAgger and a perceptual loss $\mathcal{L}_{\text{perc}}\bigl(E^{\text{real}}_{\text{per}}, E^{\text{sim}}_{\text{per}}\bigr) = \bigl\| E^{\text{real}}_{\text{per}} - E^{\text{sim}}_{\text{per}} \bigr\|_2^2$ that aligns the adapted features $E^{\text{real}}_{\text{per}} = f_{\theta_{\text{per}}^{\text{real}}}(o^{\text{real}})$ with the frozen reference $E^{\text{sim}}_{\text{per}} = f_{\theta_{\text{per}}^{\text{sim}}}(o^{\text{pix}})$. The full objective is
\[
\min_{\theta_{\text{per}}^{\text{real}}}\;
\mathbb{E}_{(o^{\text{real}}, o^{\text{pix}}) \sim \mathcal{D}_{\text{paired}}}
\Bigl[
\mathcal{L}_{\text{plan}}\bigl(\pi^{\text{real}}(o^{\text{real}}),\, \pi^{\text{sim}}(o^{\text{pix}})\bigr)
+
\lambda\, \mathcal{L}_{\text{perc}}\bigl(E^{\text{real}}_{\text{per}}, E^{\text{sim}}_{\text{per}}\bigr)
\Bigr],
\]
with $\lambda$ weighting the perceptual loss. This transfers $\pi_{\text{student}}$'s closed-loop behavior to real images using only paired observations, without human trajectory supervision.

\section{Experiments}
\label{sec:experiments}

\begin{wrapfigure}{r}{0.45\textwidth}
    \centering
    \vspace{-5mm}
    \includegraphics[width=0.43\textwidth]{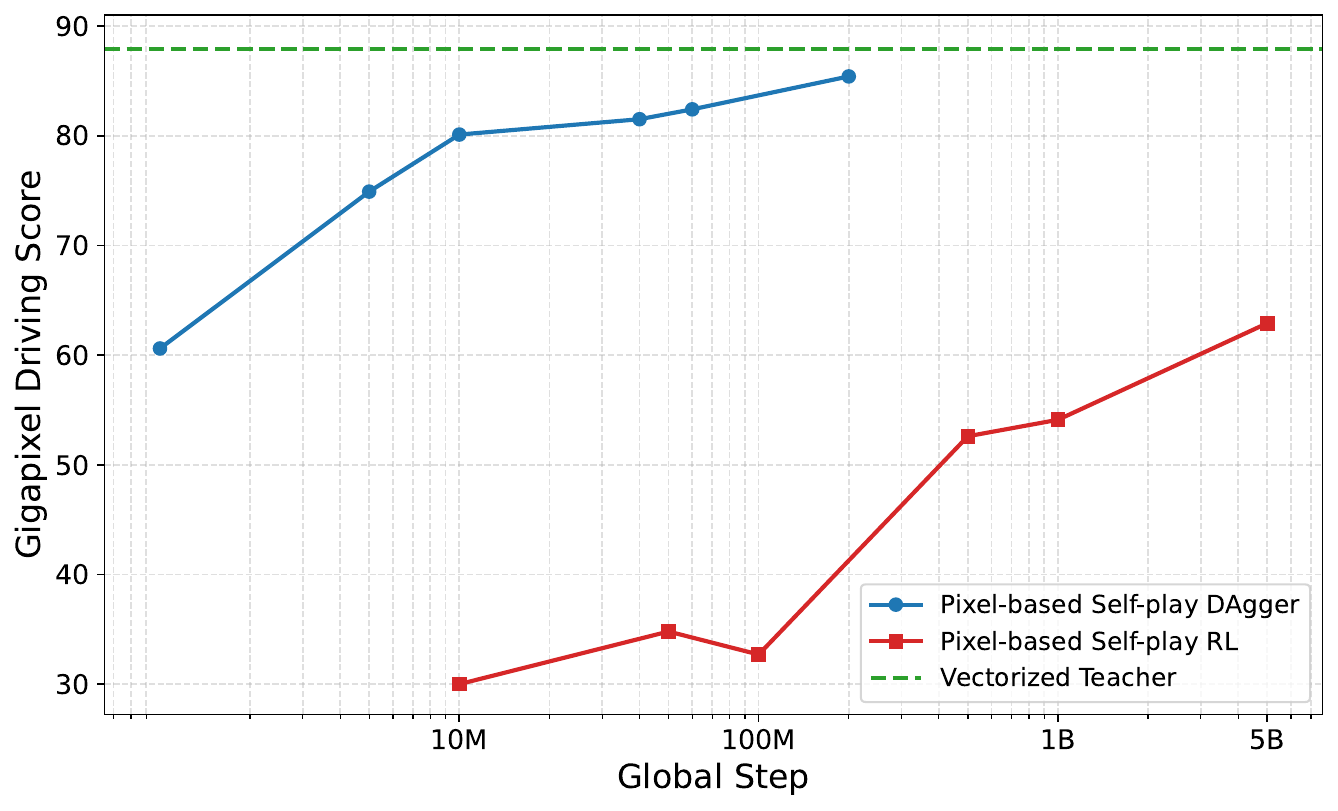}
    \caption{\textbf{Self-play DAgger vs. RL.} We compare two pixel-based training methods, self-play DAgger and self-play RL, with a CNN-based model in Gigapixel. We plot Gigapixel Driving Score ($\texttt{Completion Rate} - \texttt{Offroad Rate} - \texttt{Collision Rate}$) on a held-out set as a function of agent steps in Gigapixel. The dashed line marks the vectorized teacher performance at 25B steps.}
    \label{fig:sample-efficiency}
    \vspace{-5mm}
\end{wrapfigure}

\paragraph{Implementation Details.} We train policies in Gigapixel using 335k 20s scenarios uniformly sampled from the nuPlan train split~\cite{caesar2021nuplan}, extracting agent states, static objects, lane polylines, and traffic-light states. During self-play, vehicles are policy-controlled, while pedestrians, cyclists, and traffic lights are log-replayed. The privileged vectorized teacher $\pi_{\text{teacher}}$ follows our re-implementation of Gigaflow~\cite{cusumano2025robust}: a compact 2.7M-parameter permutation-invariant policy trained for 25B agent steps in the extracted nuPlan scenarios. We train two pixel-based student architectures: a scoring-based DrivoR model~\cite{kirby2026driving}, which predicts 64 trajectory proposals and selects the one with highest predicted PDMS, and a regression-only variant, DrivoR-Reg, which outputs a single trajectory without a scoring head. 
Both students are trained with self-play DAgger in Gigapixel for 150M steps, then adapted from simulated renderings to real NAVSIM camera images using paired simulated-real observations from \texttt{navtrain}. Further training details and hyperparameters are provided in the Appendix.

\begin{table*}[t]
\centering
\scriptsize
\resizebox{1\textwidth}{!}{
\begin{tabular}{l|c|c|c|cccc|ccccc|c|l}
 
    \toprule
    & \textbf{Human}
    & \textbf{SimScale}
    & 
    & 
    & 
    & 
    & 
    & 
    & 
    & 
    & 
    & 
    & \textbf{Per-Stage}
    &  \\
 
    \multirow{-2}{*}{\textbf{Method}} & \textbf{Traj?} & \textbf{Data?} & \multirow{-2}{*}{\textbf{Stage}} & \multirow{-2}{*}{\textbf{NC}↑} & \multirow{-2}{*}{\textbf{DAC}↑} & \multirow{-2}{*}{\textbf{DDC}↑} & \multirow{-2}{*}{\textbf{TLC}↑} & \multirow{-2}{*}{\textbf{EP}↑} & \multirow{-2}{*}{\textbf{TTC}↑} & \multirow{-2}{*}{\textbf{LK}↑} & \multirow{-2}{*}{\textbf{HC}↑} & \multirow{-2}{*}{\textbf{EC}↑} & \textbf{Score↑} & \multirow{-2}{*}{\textbf{EPDMS}↑} \\
    \midrule
 
 
     & & & S 1 & 96.2 & 79.6 & 99.1 & 99.6 & 84.1 & 95.1 & 94.2 & 97.6 & 79.1 & - & \\
    \multirow{-2}{*}{LTF~\cite{chitta2022transfuser}} & \multirow{-2}{*}{\ding{51}} & \multirow{-2}{*}{\ding{55}} & S 2 & 77.8 & 70.2 & 84.3 & 98.1 & 85.1 & 75.7 & 45.4 & 95.7 & 76.0 & - & \multirow{-2}{*}{25.1} \\
    \cmidrule{1-15}
 
     & & & S 1 & 97.1 & 94.4 & 98.8 & 99.8 & 83.9 & 96.9 & 94.7 & 96.4 & 66.2 & - & \\
    \multirow{-2}{*}{RAP~\cite{feng2025rap}} & \multirow{-2}{*}{\ding{51}} & \multirow{-2}{*}{\ding{55}} & S 2 & 83.2 & 83.9 & 87.4 & 98.0 & 86.9 & 80.4 & 52.3 & 95.2 & 52.4 & - & \multirow{-2}{*}{39.6} \\
    \cmidrule{1-15}
 
     & & & S 1 & 98.9 & 97.6 & 100.0 & 100.0 & 66.7 & 98.9 & 96.2 & 96.7 & 44.0 & - & \\
    \multirow{-2}{*}{ZTRS~\cite{li2025ztrs}} & \multirow{-2}{*}{\ding{55}} & \multirow{-2}{*}{\ding{55}} & S 2 & 91.1 & 90.4 & 95.8 & 99.0 & 63.6 & 89.8 & 60.4 & 97.6 & 66.1 & - & \multirow{-2}{*}{48.1} \\
    \cmidrule{1-15}
 
     & & & S 1 & 99.6 & 98.0 & 99.4 & 99.3 & 79.7 & 99.3 & 94.9 & 97.1 & 58.2 & - & \\
    \multirow{-2}{*}{GuideFlow~\cite{liu2025guideflow}} & \multirow{-2}{*}{\ding{51}} & \multirow{-2}{*}{\ding{55}} & S 2 & 91.4 & 89.5 & 95.2 & 98.9 & 77.5 & 89.6 & 52.6 & 93.6 & 51.0 & - & \multirow{-2}{*}{51.5} \\
    \midrule
 
    & & & S 1 & 99.6 & 99.1 & 99.9 & 100.0 & 69.6 & 99.6 & 95.8 & 95.6 & 28.4 & - & \\
    \multirow{-2}{*}{SimScale~\cite{tian2025simscale}} & \multirow{-2}{*}{\ding{51}} & \multirow{-2}{*}{\ding{51}} & S 2 & 94.5 & 94.2 & 95.8 & 99.2 & 75.8 & 92.8 & 60.1 & 96.1 & 43.2 & - & \multirow{-2}{*}{53.2} \\
    \midrule
 
    \rowcolor[HTML]{EEE5F4}
    \multicolumn{15}{c}{\textit{Regression-based (Single-Mode) DrivoR Planner}} \\
    \cmidrule{1-15}
 
    & & & S 1 & 96.1 & 76.4 & 97.6 & 99.3 & 83.3 & 95.6 & 93.1 & 97.8 & 80.4 & 65.2 & \\
    \multirow{-2}{*}{DrivoR-Reg~\cite{kirby2026driving}} & \multirow{-2}{*}{\ding{51}} & \multirow{-2}{*}{\ding{55}} & S 2 & 79.0 & 68.5 & 82.2 & 98.7 & 83.4 & 76.9 & 44.0 & 95.9 & 77.6 & 38.4 & \multirow{-2}{*}{25.5} \\
    \cmidrule{1-15}
    
    \rowcolor[rgb]{0.9,0.95,1}
     & & & S 1 & 90.2 & 78.9 & 95.3 & 99.3 & 86.4 & 88.4 & 95.8 & 97.8 & 64.4 & 60.9 & \\
    \rowcolor[rgb]{0.9,0.95,1}
    \multirow{-2}{*}{Gigapixel-DrivoR-Reg } & \multirow{-2}{*}{\ding{55}} & \multirow{-2}{*}{\ding{55}} & S 2 & 80.4 & 77.4 & 86.7 & 96.8 & 91.4 & 77.6 & 57.7 & 97.4 & 52.1 & 45.5 & \multirow{-2}{*}{29.5} \\
    \cmidrule{1-15}
 
    \rowcolor[HTML]{EEE5F4}
    \multicolumn{15}{c}{\textit{Scoring-based (Multimodal) DrivoR Planner}} \\
    \cmidrule{1-15}
 
     & & & S 1 & 98.8 & 95.1 & 98.9 & 100 & 72.6 & 98.7 & 94.0 & 97.6 & 73.3 & 80.9 & \\
    \multirow{-2}{*}{DrivoR~\cite{kirby2026driving}} & \multirow{-2}{*}{\ding{51}} & \multirow{-2}{*}{\ding{55}} & S 2 & 90.2 & 88.4 & 91.9 & 98.6 & 70.0 & 88.0 & 50.1 & 98.5 & 76.2 & 59.4 & \multirow{-2}{*}{48.3} \\
    \cmidrule{1-15}
 
     & & & S 1 & 99.1 & 98.2 & 99.3 & 99.8 & 75.4 & 98.7 & 94.9 & 97.6 & 70.2 & 84.2 & \\
    \multirow{-2}{*}{\shortstack[l]{DrivoR~\cite{kirby2026driving}\\(w/ SimScale)}} & \multirow{-2}{*}{\ding{51}} & \multirow{-2}{*}{\ding{51}} & S 2 & 92.3 & 91.6 & 97.3 & 99.1 & 75.7 & 90.6 & 56.1 & 98.4 & 44.7 & 64.6 & \multirow{-2}{*}{54.7} \\
    \cmidrule{1-15}
 
    \rowcolor[rgb]{0.9,0.95,1}
     & & & S 1 & 99.4 & 95.8 & 99.4 & 99.8 & 68.1 & 99.6 & 91.8 & 97.6 & 49.8 & 77.8 & \\
    \rowcolor[rgb]{0.9,0.95,1}
    \multirow{-2}{*}{Gigapixel-DrivoR} & \multirow{-2}{*}{\ding{55}} & \multirow{-2}{*}{\ding{55}} & S 2 & 93.7 & 92.9 & 96.0 & 98.9 & 62.6 & 90.7 & 60.2 & 98.2 & 58.8 & 63.5 & \multirow{-2}{*}{50.1} \\
 
    \bottomrule
\end{tabular}
}
\caption{\textbf{NAVSIM-v2 \texttt{navhard} Results.} Gigapixel models perform competitively across Stage 1 metrics without human trajectory supervision, and further improves Stage 2 performance, which we emphasize as the closest proxy for closed-loop execution robustness; see text for discussion.}
\label{tab:navhard}
\vspace{-6mm}
\end{table*}

\paragraph{Benchmarks and Evaluated Methods.} We evaluate in Gigapixel and on two real-world driving benchmarks: HUGSIM~\cite{zhou2025hugsim} and NAVSIM-v2~\cite{cao2025pseudo}. In Gigapixel, we run closed-loop evaluation on 1{,}000 held-out nuPlan scenarios with log-replayed surrounding actors, reporting the Gigapixel Driving Score $\max(0, \texttt{Completion Rate} - \texttt{Collision Rate} - \texttt{Off-road Rate})$. HUGSIM evaluates pixel-based policies in closed-loop on reconstructed real-world scenes using HD-Score, reported across Easy, Medium, Hard, and Extreme difficulty tiers. NAVSIM-v2 evaluates pseudo-closed-loop robustness using EPDMS on the \texttt{navhard} split; we emphasize Stage 2 EPDMS as the closest proxy for recovery under closed-loop execution, as it measures performance in perturbed ego poses. We evaluate two self-play trained policies, \textit{Gigapixel-DrivoR} and \textit{Gigapixel-DrivoR-Reg}, obtained by training DrivoR and DrivoR-Reg in Gigapixel with self-play DAgger followed by sim-to-real perception adaptation. We compare against behavior-cloned DrivoR variants, DrivoR with SimScale recovery data, published leaderboard methods, and ablations that replace self-play DAgger with behavior cloning or single-agent DAgger. Full benchmark protocols and metrics, baseline training details, and metric definitions are provided in the Appendix.

\paragraph{Results.} Figure~\ref{fig:training_sps} compares Gigapixel's rendering throughput against alternative pixel renderers. The Gigapixel rasterizer is $\sim1000\times$ faster than the Gaussian-splatting-based HUGSIM \cite{zhou2025hugsim} and $\sim4000\times$ faster than the rasterization-based RAP \cite{feng2025rap} at the $512\times512$ resolution. Although RAP adopts a similar simplified scene representation, it runs entirely on CPU and forgoes the batched entity-component-system (ECS) design that underpins Madrona's high throughput \cite{shacklett2023extensible}. Figure~\ref{fig:sample-efficiency} compares self-play DAgger training against direct self-play RL training from pixels for an action-based CNN policy in Gigapixel. We use the CNN policy as it is a lighter weight architecture than DrivoR, thus enabling higher throughput training for RL experimentation (see Figure~\ref{fig:training_sps}). Self-play DAgger is dramatically more sample-efficient: it surpasses a Gigapixel Driving Score of 60 in roughly $3000\times$ fewer steps than self-play RL, motivating our DAgger-based approach.

\begin{table}
    \centering
    \small
    \setlength{\tabcolsep}{3pt}
    \begin{tabular}{@{}lcc|ccccc@{}}
    \toprule
    Gigapixel & Perception & Frozen & \multicolumn{5}{c}{HD-Score} \\
    Training Strategy & L2 Loss & Planning Head? & E & M & H & X & Avg. \\
    \midrule
    Self-play DAgger & \ding{55} & \ding{55} &
    43.5 & 10.5 & 7.3 & 16.4 & 15.8 \\
    \cmidrule{1-8}
    Self-play DAgger & \ding{55} & \ding{51} &
    52.6 & 15.1 & 7.2 & 15.2 & 18.5 \\
    \midrule
    \midrule
    DAgger & \ding{51} & \ding{51} &
    61.8 & 32.0 & 15.1 & 24.3 & \underline{30.1} \\
    \cmidrule{1-8}
    BC & \ding{51} & \ding{51} &
    45.2 & 22.7 & 10.9 & 4.9 & 18.6 \\
    \cmidrule{1-8}
    Self-play DAgger & \ding{51} & \ding{51} &
    65.8 & 38.8 & 13.1 & 24.6 & \textbf{33.2} \\
    \bottomrule
    \end{tabular}
    \vspace{3mm}
    \caption{\textbf{Ablation study.} We ablate the Gigapixel training strategy, and perception adaptation design on HUGSIM. All variants use DrivoR-Reg. Best method is \textbf{bolded}; second-best \underline{underlined}.}
    \vspace{-8mm}
    \label{tab:ablation}
\end{table}

\begin{wrapfigure}{r}{0.45\textwidth}
    \centering
    \vspace{-5mm}
    \includegraphics[width=0.43\textwidth]{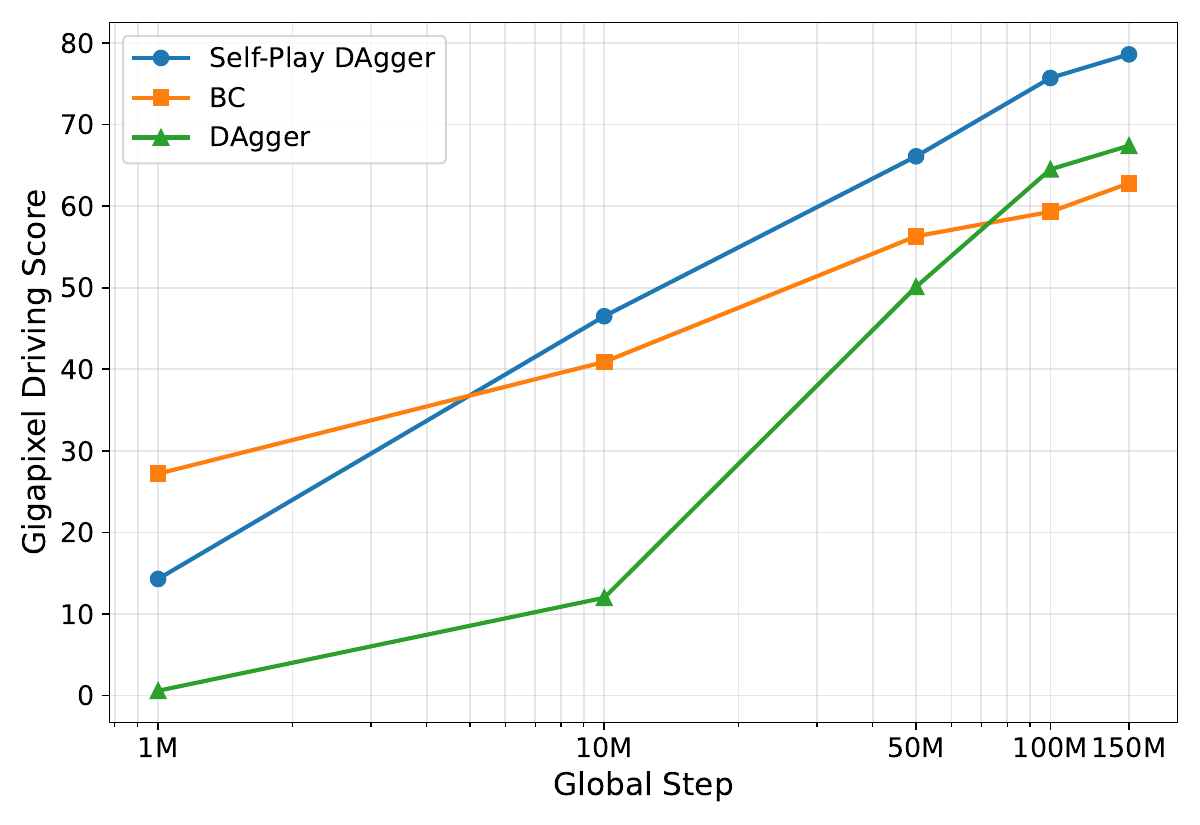}
    \caption{\textbf{Scaling Experiments.} We compare different end-to-end training strategies at scale in Gigapixel. We plot Gigapixel Driving Score ($\texttt{Completion Rate} - \texttt{Offroad Rate} - \texttt{Collision Rate}$) on a held-out set as a function of agent steps in Gigapixel. In all configurations, we train the single-mode DrivoR-Reg architecture.}
    \label{fig:gigapixel_driving_score}
    \vspace{-3mm}
\end{wrapfigure}

Table~\ref{tab:hugsim} reports closed-loop driving performance on the HUGSIM benchmark. Gigapixel-DrivoR achieves state-of-the-art performance, outperforming all baselines on both average RC (50.1) and HD-Score (38.5)---without any human trajectory supervision. The scoring-based Gigapixel-DrivoR improves over its behavior-cloning counterpart DrivoR by 2.8 HD-Score points (38.5 vs.~35.7), and the regression-based Gigapixel-DrivoR-Reg improves over DrivoR-Reg by 12.5 points (33.2 vs.~20.7)---a 60\% relative gain that demonstrates the impact of closed-loop self-play DAgger training. The one regime where DrivoR exceeds Gigapixel-DrivoR is the Extreme tier (32.5 vs.~21.6 HD-Score), where surrounding actors are most adversarial. Manual inspection (examples in the supplementary) reveals a \textit{high-velocity bias} in DrivoR: it tends to drive fast, which incidentally allows it to outpace adversarial actors attempting to induce collisions. Gigapixel-DrivoR drives more cautiously and instead yields to these actors, often getting stuck. To quantify this, we measured the average collision velocity across the full evaluation set: DrivoR collides at $5.27$ m/s on average, compared to $1.95$ m/s for Gigapixel-DrivoR---a $2.7\times$ reduction, indicating that DrivoR's Extreme-tier advantage stems from a less safe driving style rather than better closed-loop reasoning.

\begin{figure*}[t]
    \centering
    \includegraphics[width=\textwidth]{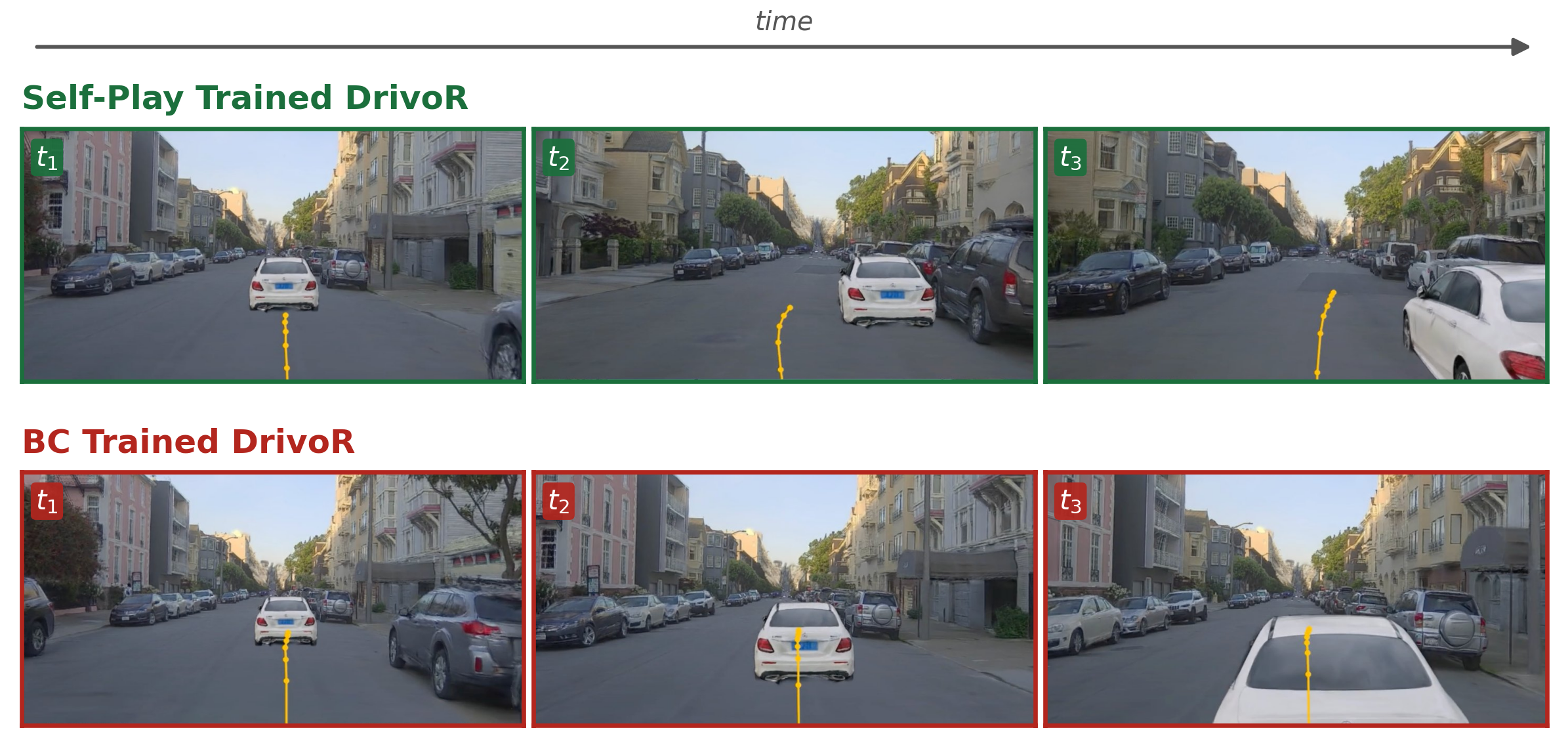}
    \caption{\textbf{Qualitative comparison of closed-loop driving behavior near a
    decelerating lead vehicle.} Frames advance left to right ($t_1 \rightarrow t_3$);
    the yellow waypoints denote the model's planned future trajectory. \textbf{Top
    (green): Self-Play Trained DrivoR.} As the lead vehicle comes to a stop, the policy
    reduces speed and plans a smooth lateral trajectory that nudges out of the lane and
    passes the stationary vehicle safely. \textbf{Bottom (red): BC Trained DrivoR.} The
    behavior-cloned policy keeps a centered
    straight-ahead plan, closes distance on the lead vehicle, and ultimately rear-ends it.}
    \vspace{-5mm}
    \label{fig:drivor_qualitative}
\end{figure*}

Table~\ref{tab:navhard} reports pseudo-closed-loop performance on NAVSIM-v2 \texttt{navhard}. Without human trajectory supervision or SimScale data, Gigapixel-DrivoR and Gigapixel-DrivoR-Reg outperform their behavior-cloning counterparts in EPDMS (50.1 vs.~48.3 and 29.5 vs.~25.5, respectively). These gains are concentrated in Stage 2, which evaluates planning from perturbed off-distribution ego poses and is therefore the closest proxy for closed-loop robustness: Gigapixel-DrivoR-Reg improves over DrivoR-Reg by 7.1 points (45.5 vs.~38.4), while Gigapixel-DrivoR improves over DrivoR by 4.1 points (63.5 vs.~59.4). DrivoR (w/ SimScale) achieves the best overall EPDMS (54.7), but SimScale's data curation explicitly mines high-EPDMS trajectories, whereas our Gigaflow teacher is not trained with a NAVSIM-v2-specific objective. Even so, Gigapixel-DrivoR nearly matches its Stage 2 score (63.5 vs.~64.6).

Figure~\ref{fig:gigapixel_driving_score} reports closed-loop performance of the DrivoR-Reg student across training strategies as experience scales. Self-play DAgger improves consistently with scale and outperforms both single-agent DAgger and behavior cloning (BC) beyond 10M steps. BC plateaus around 100M steps, since the student is never exposed to states from its own rollouts. Self-play DAgger also outperforms single-agent DAgger throughout, reflecting the two factors identified in Section~\ref{subsec:self-play-dagger}: every agent in a rollout contributes supervised data, and the resulting interactions span more diverse multi-agent behaviors. Table~\ref{tab:ablation} confirms these gains transfer after sim-to-real perception adaptation on the HUGSIM benchmark: self-play DAgger achieves a 3.1-point HD-Score improvement over single-agent DAgger and 14.6 points over BC, mirroring the in-simulation ranking. The table also ablates our adaptation design: removing the perceptual loss drops HD-Score from 33.2 to 18.5, and additionally unfreezing the planning head drops it to 15.8---confirming that both choices are essential for preserving the closed-loop behavior learned during self-play.

Figure~\ref{fig:drivor_qualitative} contrasts the two policies as the ego vehicle approaches
a lead vehicle slowing to a stop. The self-play trained DrivoR (top row) anticipates
the deceleration, reduces speed, and once the lead vehicle stops plans a safe collision-free lateral
maneuver around it. The BC trained DrivoR (bottom row),
by contrast, maintains a high-velocity straight-ahead trajectory ($t_1 \rightarrow t_3$) and
ultimately rear-ends the stopped vehicle. The failure exposes a limitation of behavior cloning: the
rare, safety-critical states that precede a stop are underrepresented at training time, so
the policy never learns the deceleration-and-avoid response that self-play discovers through explicit closed-loop interaction.


\begin{figure*}[t]
    \centering
    \includegraphics[width=\textwidth]{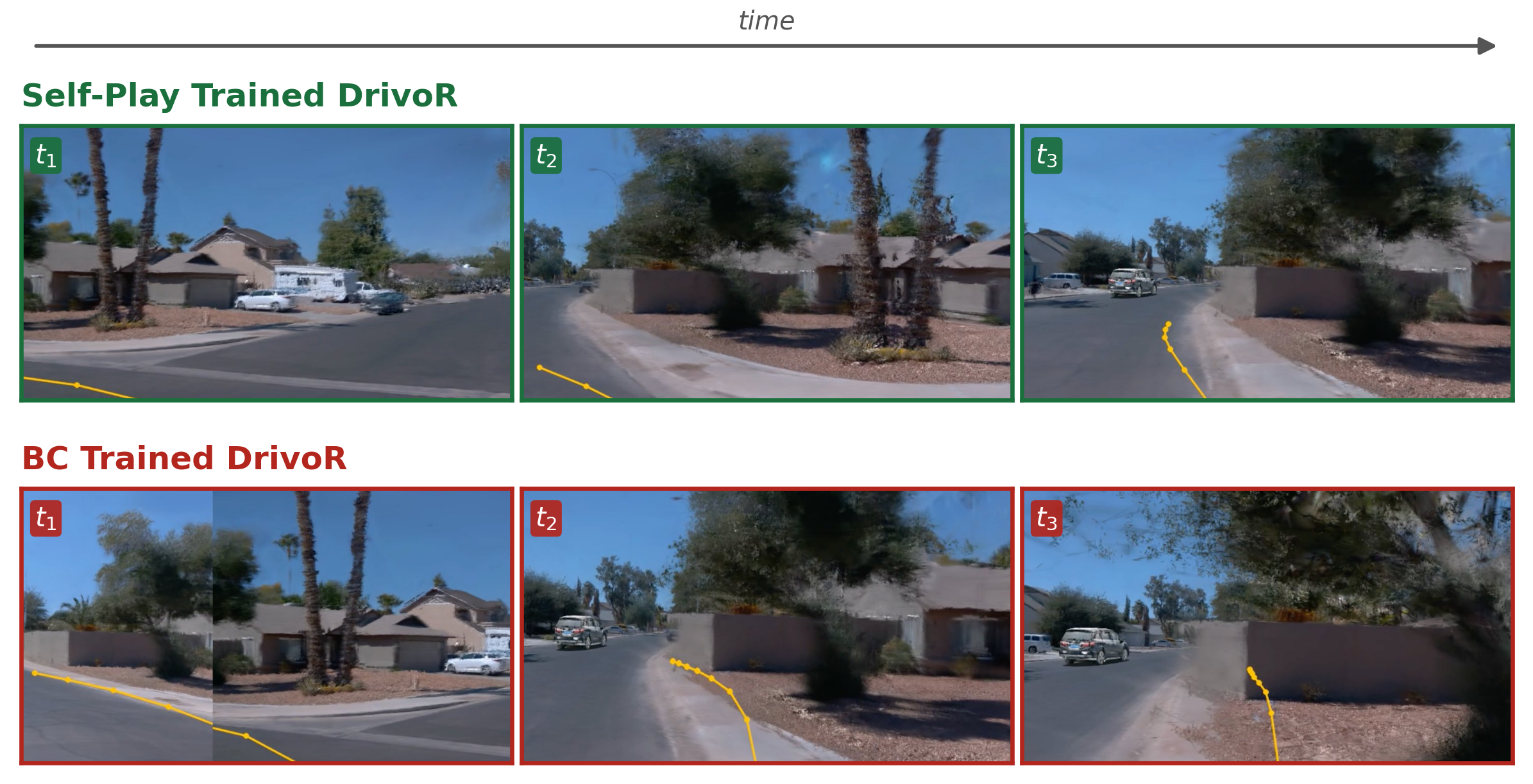}
    \caption{\textbf{Behavior from a recovery state near the road edge.} Frames advance in
    time ($t_1 \rightarrow t_3$); yellow waypoints denote the planned trajectory. \textbf{Top
    (green): Self-Play Trained DrivoR} plans a slow, careful turn that corrects toward the
    lane and keeps the vehicle on the drivable surface. \textbf{Bottom (red): BC Trained
    DrivoR} fails to correct adequately near the road edge and drifts off-road.}
    \label{fig:drivor_recovery_compare}
    \vspace{-4mm}
\end{figure*}

Figure~\ref{fig:drivor_recovery_compare} compares the two policies initialized in a recovery
state, where the vehicle starts near the road boundary and a slow, careful turn is required
to avoid going off-road. The self-play trained DrivoR (top row) reduces speed and
plans a corrective trajectory, recovering safely to a nominal lane
position. The BC trained DrivoR (bottom row) does not steer enough near the edge,
and its trajectory drifts off the drivable surface, resulting in an off-road violation. Such
corrective states are frequently visited during self-play training, but are rarely present in offline
expert logs.
\section{Conclusion, Limitations, and Future Work}


We present the first method for effectively training end-to-end driving policies via self-play. Our approach combines Gigapixel for high-throughput pixel-based simulation, self-play DAgger for distilling a privileged RL teacher, and lightweight sim-to-real adaptation, achieving state-of-the-art HUGSIM performance and strong NAVSIM-v2 performance. These contributions establish self-play as a viable alternative to large-scale behavior cloning for training end-to-end driving models.


\paragraph{Limitations.} Gigapixel is limited by its abstract scene representation: agent boxes, lane polylines, and traffic-light states cannot capture cues such as debris, unusual obstacles, weather, or lighting. It also introduces a teacher--student asymmetry \cite{nguyen2026lead}, where the teacher observes privileged vectorized observations $o^{\text{vec}}$ while the student must infer actions from pixels $o^{\text{pix}}$, which can be unrecoverable under occlusion or limited visibility. Finally, sim-to-real adaptation requires paired data $\mathcal{D}_{\text{paired}}$, which is practical on NAVSIM but less direct with only raw sensor logs.


\paragraph{Future Work.} Several directions follow naturally from this work. First, while we deliberately forgo human trajectory supervision to demonstrate the strength of self-play alone, the most capable real-world policies will likely combine both signals. Human data could be incorporated through KL-regularization during self-play training~\cite{cornelissehuman} or through post-training toward human-like behavior. Second, Gigapixel currently initializes scenarios from nuPlan logs, which constrains the distribution of initial scene configurations. Generative approaches~\cite{chitta2024sledge, rowe2025scenario} could synthesize initial conditions that induce rare or informative interactions, further enriching the self-play curriculum. Finally, training pixel-based RL policies directly in Gigapixel with native trajectory outputs remains an open challenge, but would eliminate the teacher-student asymmetry inherent in distillation approaches.


\clearpage
\acknowledgments{We thank Samsung, the IVADO and the Canada First
Research Excellence Fund (CFREF) / Apogee Funds,
the Canada CIFAR AI Chairs Program, Fonds de Recherche Nature et Technologies (FRQNT), and the NSERC
Discovery Grants and Doctoral Scholarship programs for their financial support. We thank Mila - Quebec AI Institute for compute resources. Daphne Cornelisse is partially supported by the Cooperative AI Foundation and a Chishiki-AI SCIPE Fellowship.}


\bibliography{main}  

@String(ECCV  = {Eur. Conf. Comput. Vis.})

@String(ICLR  = {Int. Conf. Learn. Represent.})

@String(JMLR  = {J. Mach. Learn. Res.})

@String(TOG   = {ACM Trans. Graph.})

@String(ECCV  = {ECCV})

@String(ICLR  = {ICLR})

@String(JMLR  = {JMLR})

@String(TOG   = {ACM TOG})

@article{chen2024end,
  title={End-to-end autonomous driving: Challenges and frontiers},
  author={Chen, Li and Wu, Penghao and Chitta, Kashyap and Jaeger, Bernhard and Geiger, Andreas and Li, Hongyang},
  journal={IEEE Transactions on Pattern Analysis and Machine Intelligence},
  volume={46},
  number={12},
  pages={10164--10183},
  year={2024},
  publisher={IEEE}
}

@inproceedings{hu2023planning,
  title={Planning-oriented autonomous driving},
  author={Hu, Yihan and Yang, Jiazhi and Chen, Li and Li, Keyu and Sima, Chonghao and Zhu, Xizhou and Chai, Siqi and Du, Senyao and Lin, Tianwei and Wang, Wenhai and others},
  booktitle={Proceedings of the IEEE/CVF conference on computer vision and pattern recognition},
  pages={17853--17862},
  year={2023}
}

@inproceedings{jiang2023vad,
  title={Vad: Vectorized scene representation for efficient autonomous driving},
  author={Jiang, Bo and Chen, Shaoyu and Xu, Qing and Liao, Bencheng and Chen, Jiajie and Zhou, Helong and Zhang, Qian and Liu, Wenyu and Huang, Chang and Wang, Xinggang},
  booktitle={Proceedings of the IEEE/CVF International Conference on Computer Vision},
  pages={8340--8350},
  year={2023}
}

@article{chitta2022transfuser,
  title={Transfuser: Imitation with transformer-based sensor fusion for autonomous driving},
  author={Chitta, Kashyap and Prakash, Aditya and Jaeger, Bernhard and Yu, Zehao and Renz, Katrin and Geiger, Andreas},
  journal={IEEE transactions on pattern analysis and machine intelligence},
  volume={45},
  number={11},
  pages={12878--12895},
  year={2022},
  publisher={IEEE}
}

@article{hwang2024emma,
  title={Emma: End-to-end multimodal model for autonomous driving},
  author={Hwang, Jyh-Jing and Xu, Runsheng and Lin, Hubert and Hung, Wei-Chih and Ji, Jingwei and Choi, Kristy and Huang, Di and He, Tong and Covington, Paul and Sapp, Benjamin and others},
  journal={arXiv preprint arXiv:2410.23262},
  year={2024}
}

@article{rowe2025poutine,
  title={Poutine: Vision-language-trajectory pre-training and reinforcement learning post-training enable robust end-to-end autonomous driving},
  author={Rowe, Luke and de Schaetzen, Rodrigue and Girgis, Roger and Pal, Christopher and Paull, Liam},
  journal={arXiv preprint arXiv:2506.11234},
  year={2025}
}

@inproceedings{naumann2025data,
  title={Data scaling laws for end-to-end autonomous driving},
  author={Naumann, Alexander and Gu, Xunjiang and Dimlioglu, Tolga and Bojarski, Mariusz and Degirmenci, Alperen and Popov, Alexander and Bisla, Devansh and Pavone, Marco and Muller, Urs and Ivanovic, Boris},
  booktitle={Proceedings of the Computer Vision and Pattern Recognition Conference},
  pages={2571--2582},
  year={2025}
}

@article{baniodeh2025scaling,
  title={Scaling Laws of Motion Forecasting and Planning--Technical Report},
  author={Baniodeh, Mustafa and Goel, Kratarth and Ettinger, Scott and Fuertes, Carlos and Seff, Ari and Shen, Tim and Gulino, Cole and Yang, Chenjie and Jerfel, Ghassen and Choe, Dokook and others},
  journal={arXiv preprint arXiv:2506.08228},
  year={2025}
}

@inproceedings{codevilla2019exploring,
  title={Exploring the limitations of behavior cloning for autonomous driving},
  author={Codevilla, Felipe and Santana, Eder and L{\'o}pez, Antonio M and Gaidon, Adrien},
  booktitle={Proceedings of the IEEE/CVF international conference on computer vision},
  pages={9329--9338},
  year={2019}
}

@inproceedings{ross2011reduction,
  title={A reduction of imitation learning and structured prediction to no-regret online learning},
  author={Ross, St{\'e}phane and Gordon, Geoffrey and Bagnell, Drew},
  booktitle={Proceedings of the fourteenth international conference on artificial intelligence and statistics},
  pages={627--635},
  year={2011},
  organization={JMLR Workshop and Conference Proceedings}
}

@article{karkus2025beyond,
  title={Beyond Behavior Cloning in Autonomous Driving: a Survey of Closed-Loop Training Techniques},
  author={Karkus, Peter and Igl, Maximilian and Chen, Yuxiao and Chitta, Kashyap and Packer, Jef and Douillard, Bertrand and Tian, Ran and Naumann, Alexander and Garcia-Cobo, Guillermo and Tan, Shuhan and others},
  journal={Authorea Preprints},
  year={2025},
  publisher={Authorea}
}

@inproceedings{cusumano2025robust,
  title={Robust Autonomy Emerges from Self-Play},
  author={Cusumano-Towner, Marco and Hafner, David and Hertzberg, Alexander and Huval, Brody and Petrenko, Aleksei and Vinitsky, Eugene and Wijmans, Erik and Killian, Taylor W and Bowers, Stuart and Sener, Ozan and others},
  booktitle={International Conference on Machine Learning},
  pages={11710--11737},
  year={2025},
  organization={PMLR}
}

@article{zheng2024data,
  title={Data Scaling Laws for Imitation Learning-Based End-to-End Autonomous Driving},
  author={Zheng, Yupeng and Yang, Pengxuan and Xia, Zhongpu and Zhang, Qichao and Zheng, Yuhang and Gu, Songen and Jin, Bu and Zhang, Teng and Lu, Ben and Han, Chao and others},
  journal={arXiv preprint arXiv:2412.02689},
  year={2024}
}

@article{chib2023recent,
  title={Recent advancements in end-to-end autonomous driving using deep learning: A survey},
  author={Chib, Pranav Singh and Singh, Pravendra},
  journal={IEEE Transactions on Intelligent Vehicles},
  volume={9},
  number={1},
  pages={103--118},
  year={2023},
  publisher={IEEE}
}

@software{pufferdrive2025github,
  author = {Daphne Cornelisse and Spencer Cheng and Pragnay Mandavilli and Julian Hunt and Kevin Joseph and Waël Doulazmi and Valentin Charraut and Aditya Gupta and Joseph Suarez and Eugene Vinitsky},
  title = {{PufferDrive}: A Fast and Friendly Driving Simulator for Training and Evaluating {RL} Agents},
  url = {https://github.com/Emerge-Lab/PufferDrive},
  version = {2.0.0},
  year = {2025},
}

@inproceedings{zhang2024asymmetric,
  title={Learning to drive via asymmetric self-play},
  author={Zhang, Chris and Biswas, Sourav and Wong, Kelvin and Fallah, Kion and Zhang, Lunjun and Chen, Dian and Casas, Sergio and Urtasun, Raquel},
  booktitle={European Conference on Computer Vision},
  pages={149--168},
  year={2024},
  organization={Springer}
}

@inproceedings{cornelissehuman,
  title={Human-compatible driving agents through data-regularized self-play reinforcement learning},
  author={Cornelisse, Daphne and Vinitsky, Eugene},
  booktitle={Reinforcement Learning Conference},
  year={2024}
}

@article{cornelisse2025building,
  title={Building reliable sim driving agents by scaling self-play},
  author={Cornelisse, Daphne and Pandya, Aarav and Joseph, Kevin and Su{\'a}rez, Joseph and Vinitsky, Eugene},
  journal={arXiv preprint arXiv:2502.14706},
  year={2025}
}

@inproceedings{rosenzweig2024high,
  title={High-throughput batch rendering for embodied ai},
  author={Rosenzweig, Luc Guy and Shacklett, Brennan and Xia, Warren and Fatahalian, Kayvon},
  booktitle={SIGGRAPH Asia 2024 Conference Papers},
  pages={1--9},
  year={2024}
}

@inproceedings{agarwal2024policy,
  title={On-policy distillation of language models: Learning from self-generated mistakes},
  author={Agarwal, Rishabh and Vieillard, Nino and Zhou, Yongchao and Stanczyk, Piotr and Garea, Sabela Ramos and Geist, Matthieu and Bachem, Olivier},
  booktitle={The twelfth international conference on learning representations},
  year={2024}
}

@article{zhou2025hugsim,
  title={Hugsim: A real-time, photo-realistic and closed-loop simulator for autonomous driving},
  author={Zhou, Hongyu and Lin, Longzhong and Wang, Jiabao and Lu, Yichong and Bai, Dongfeng and Liu, Bingbing and Wang, Yue and Geiger, Andreas and Liao, Yiyi},
  journal={IEEE Transactions on Pattern Analysis and Machine Intelligence},
  year={2025},
  publisher={IEEE}
}

@inproceedings{cao2025pseudo,
  title={Pseudo-Simulation for Autonomous Driving},
  author={Cao, Wei and Hallgarten, Marcel and Li, Tianyu and Dauner, Daniel and Gu, Xunjiang and Wang, Caojun and Miron, Yakov and Aiello, Marco and Li, Hongyang and Gilitschenski, Igor and others},
  booktitle={Conference on Robot Learning},
  pages={4709--4722},
  year={2025},
  organization={PMLR}
}

@article{vinitsky2022nocturne,
  title={Nocturne: a scalable driving benchmark for bringing multi-agent learning one step closer to the real world},
  author={Vinitsky, Eugene and Lichtl{\'e}, Nathan and Yang, Xiaomeng and Amos, Brandon and Foerster, Jakob},
  journal={Advances in Neural Information Processing Systems},
  volume={35},
  pages={3962--3974},
  year={2022}
}

@article{gulino2024waymax,
  title={Waymax: An accelerated, data-driven simulator for large-scale autonomous driving research},
  author={Gulino, Cole and Fu, Justin and Luo, Wenjie and Tucker, George and Bronstein, Eli and Lu, Yiren and Harb, Jean and Pan, Xinlei and Wang, Yan and Chen, Xiangyu and others},
  journal={Advances in Neural Information Processing Systems},
  volume={36},
  year={2023}
}

@inproceedings{gpudrive2025,
  title={{GPUDrive}: Data-driven, multi-agent driving simulation at 1 million {FPS}},
  author={Kazemkhani, Saman and Pandya, Aarav and Cornelisse, Daphne and Shacklett, Brennan and Vinitsky, Eugene},
  booktitle={The Thirteenth International Conference on Learning Representations, {ICLR} 2025},
  year={2025}
}

@inproceedings{dosovitskiy2017carla,
  title={{CARLA}: An open urban driving simulator},
  author={Dosovitskiy, Alexey and Ros, German and Codevilla, Felipe and Lopez, Antonio and Koltun, Vladlen},
  booktitle={Conference on Robot Learning},
  pages={1--16},
  year={2017},
  organization={PMLR}
}

@article{yang2025resim,
  title={{ReSim}: Reliable World Simulation for Autonomous Driving},
  author={Yang, Jiazhi and Chitta, Kashyap and Gao, Shenyuan and Chen, Long and Shao, Yuqian and Jia, Xiaosong and Li, Hongyang and Geiger, Andreas and Yue, Xiangyu and Chen, Li},
  journal={arXiv preprint arXiv:2506.09981},
  year={2025}
}

@inproceedings{
feng2025rap,
title={{RAP}: 3D Rasterization Augmented End-to-End Planning},
author={Lan Feng and Yang Gao and Eloi Zablocki and Quanyi Li and Wuyang Li and Sichao Liu and Matthieu Cord and Alexandre Alahi},
booktitle={The Fourteenth International Conference on Learning Representations},
year={2026},
url={https://openreview.net/forum?id=a9bOgeqbdB}
}

@article{suarez2024pufferlib,
  title={{PufferLib}: Making Reinforcement Learning Libraries and Environments Play Nice},
  author={Suarez, Joseph},
  journal={arXiv preprint arXiv:2406.12905},
  year={2024}
}

@article{shacklett2023extensible,
  title={An extensible, data-oriented architecture for high-performance, many-world simulation},
  author={Shacklett, Brennan and Rosenzweig, Luc Guy and Xie, Zhiqiang and Sarkar, Bidipta and Szot, Andrew and Wijmans, Erik and Koltun, Vladlen and Batra, Dhruv and Fatahalian, Kayvon},
  journal={ACM Transactions on Graphics (TOG)},
  volume={42},
  number={4},
  pages={1--13},
  year={2023},
  publisher={ACM New York, NY, USA}
}

@article{silver2017alphazero,
  title={Mastering Chess and Shogi by Self-Play with a General Reinforcement Learning Algorithm},
  author={Silver, David and Hubert, Thomas and Schrittwieser, Julian and Antonoglou, Ioannis and Lai, Matthew and Guez, Arthur and Lanctot, Marc and Sifre, Laurent and Kumaran, Dharshan and Graepel, Thore and Lillicrap, Timothy and Simonyan, Karen and Hassabis, Demis},
  journal={arXiv preprint arXiv:1712.01815},
  year={2017}
}

@article{go2017,
  title={Mastering the game of {Go} without human knowledge},
  author={Silver, David and Schrittwieser, Julian and Simonyan, Karen and Antonoglou, Ioannis and Huang, Aja and Guez, Arthur and Hubert, Thomas and Baker, Lucas and Lai, Matthew and Bolton, Adrian and Chen, Yutian and Lillicrap, Timothy and Hui, Fan and Sifre, Laurent and van den Driessche, George and Graepel, Thore and Hassabis, Demis},
  journal={Nature},
  volume={550},
  number={7676},
  pages={354--359},
  year={2017}
}

@article{population2018,
  title={Human-level performance in first-person multiplayer games with population-based deep reinforcement learning},
  author={Jaderberg, Max and Czarnecki, Wojciech M. and Dunning, Iain and Marris, Luke and Lever, Guy and Garc{\'i}a Casta{\~n}eda, Antonio and Beattie, Charles and Rabinowitz, Neil C. and Morcos, Ari S. and Ruderman, Avraham and Sonnerat, Nicolas and Green, Tim and Deason, Louise and Leibo, Joel Z. and Silver, David and Hassabis, Demis and Kavukcuoglu, Koray and Graepel, Thore},
  journal={arXiv preprint arXiv:1807.01281},
  year={2018}
}

@inproceedings{diplomacy2023,
  title={Mastering the Game of No-Press Diplomacy via Human-Regularized Reinforcement Learning and Planning},
  author={Bakhtin, Anton and Wu, David J. and Lerer, Adam and Gray, Jonathan and Jacob, Athul Paul and Farina, Gabriele and Miller, Alexander H. and Brown, Noam},
  booktitle={The Eleventh International Conference on Learning Representations, {ICLR} 2023},
  year={2023}
}

@article{wang2026nomad,
  title={Learning to Drive in New Cities Without Human Demonstrations},
  author={Wang, Zilin and Rahmani, Saeed and Cornelisse, Daphne and Sarkar, Bidipta and Goldie, Alexander David and Foerster, Jakob Nicolaus and Whiteson, Shimon},
  journal={arXiv preprint arXiv:2602.15891},
  year={2026}
}

@inproceedings{dauner2023pdmclosed,
  title={Parting with misconceptions about learning-based vehicle motion planning},
  author={Dauner, Daniel and Hallgarten, Marcel and Geiger, Andreas and Chitta, Kashyap},
  booktitle={Conference on Robot Learning},
  pages={1268--1281},
  year={2023},
  organization={PMLR}
}

@article{li2025ztrs,
  title={Ztrs: Zero-imitation end-to-end autonomous driving with trajectory scoring},
  author={Li, Zhenxin and Yao, Wenhao and Wang, Zi and Sun, Xinglong and Chen, Jingde and Chang, Nadine and Shen, Maying and Song, Jingyu and Wu, Zuxuan and Lan, Shiyi and others},
  journal={arXiv preprint arXiv:2510.24108},
  year={2025}
}

@article{liu2025guideflow,
  title={GuideFlow: Constraint-guided flow matching for planning in end-to-end autonomous driving},
  author={Liu, Lin and Jia, Caiyan and Yu, Guanyi and Song, Ziying and Li, JunQiao and Jia, Feiyang and Wu, Peiliang and Hao, Xiaoshuai and Luo, Yadan},
  journal={arXiv preprint arXiv:2511.18729},
  year={2025}
}

@article{tian2025simscale,
  title={Simscale: Learning to drive via real-world simulation at scale},
  author={Tian, Haochen and Li, Tianyu and Liu, Haochen and Yang, Jiazhi and Qiu, Yihang and Li, Guang and Wang, Junli and Gao, Yinfeng and Zhang, Zhang and Wang, Liang and others},
  journal={arXiv preprint arXiv:2511.23369},
  year={2025}
}

@article{kirby2026driving,
  title={Driving on Registers},
  author={Kirby, Ellington and Boulch, Alexandre and Xu, Yihong and Yin, Yuan and Puy, Gilles and Zablocki, {\'E}loi and Bursuc, Andrei and Gidaris, Spyros and Marlet, Renaud and Bartoccioni, Florent and others},
  journal={arXiv preprint arXiv:2601.05083},
  year={2026}
}

@article{ljungbergh2024neuroncap,
  title={NeuroNCAP: Photorealistic Closed-loop Safety Testing for Autonomous Driving},
  author={Ljungbergh, William and Tonderski, Adam and Johnander, Joakim and Caesar, Holger and {\AA}str{\"o}m, Kalle and Felsberg, Michael and Petersson, Christoffer},
  journal={European Conference on Computer Vision (ECCV)},
  year={2024}
}

@inproceedings{sun2020scalability,
  title={Scalability in perception for autonomous driving: Waymo open dataset},
  author={Sun, Pei and Kretzschmar, Henrik and Dotiwalla, Xerxes and Chouard, Aurelien and Patnaik, Vijaysai and Tsui, Paul and Guo, James and Zhou, Yin and Chai, Yuning and Caine, Benjamin and others},
  booktitle={Proceedings of the IEEE/CVF conference on computer vision and pattern recognition},
  pages={2446--2454},
  year={2020}
}

@inproceedings{caesar2020nuscenes,
  title={nuscenes: A multimodal dataset for autonomous driving},
  author={Caesar, Holger and Bankiti, Varun and Lang, Alex H and Vora, Sourabh and Liong, Venice Erin and Xu, Qiang and Krishnan, Anush and Pan, Yu and Baldan, Giancarlo and Beijbom, Oscar},
  booktitle={Proceedings of the IEEE/CVF conference on computer vision and pattern recognition},
  pages={11621--11631},
  year={2020}
}

@article{caesar2021nuplan,
  title={nuplan: A closed-loop ml-based planning benchmark for autonomous vehicles},
  author={Caesar, Holger and Kabzan, Juraj and Tan, Kok Seang and Fong, Whye Kit and Wolff, Eric and Lang, Alex and Fletcher, Luke and Beijbom, Oscar and Omari, Sammy},
  journal={arXiv preprint arXiv:2106.11810},
  year={2021}
}

@article{xu2025wod,
  title={Wod-e2e: Waymo open dataset for end-to-end driving in challenging long-tail scenarios},
  author={Xu, Runsheng and Lin, Hubert and Jeon, Wonseok and Feng, Hao and Zou, Yuliang and Sun, Liting and Gorman, John and Tolstaya, Ekaterina and Tang, Sarah and White, Brandyn and others},
  journal={arXiv preprint arXiv:2510.26125},
  year={2025}
}

@article{ghilotti2026truckdrive,
  title={TruckDrive: Long-Range Autonomous Highway Driving Dataset},
  author={Ghilotti, Filippo and Palladin, Edoardo and Brucker, Samuel and Sigal, Adam and Bijelic, Mario and Heide, Felix},
  journal={arXiv preprint arXiv:2603.02413},
  year={2026}
}

@inproceedings{arai2025covla,
  title={Covla: Comprehensive vision-language-action dataset for autonomous driving},
  author={Arai, Hidehisa and Miwa, Keita and Sasaki, Kento and Watanabe, Kohei and Yamaguchi, Yu and Aoki, Shunsuke and Yamamoto, Issei},
  booktitle={2025 IEEE/CVF Winter Conference on Applications of Computer Vision (WACV)},
  pages={1933--1943},
  year={2025},
  organization={IEEE}
}

@inproceedings{weng2024drive,
  title={Para-drive: Parallelized architecture for real-time autonomous driving},
  author={Weng, Xinshuo and Ivanovic, Boris and Wang, Yan and Wang, Yue and Pavone, Marco},
  booktitle={Proceedings of the IEEE/CVF Conference on Computer Vision and Pattern Recognition},
  pages={15449--15458},
  year={2024}
}

@article{li2024hydra,
  title={Hydra-mdp: End-to-end multimodal planning with multi-target hydra-distillation},
  author={Li, Zhenxin and Li, Kailin and Wang, Shihao and Lan, Shiyi and Yu, Zhiding and Ji, Yishen and Li, Zhiqi and Zhu, Ziyue and Kautz, Jan and Wu, Zuxuan and others},
  journal={arXiv preprint arXiv:2406.06978},
  year={2024}
}

@inproceedings{liao2025diffusiondrive,
  title={Diffusiondrive: Truncated diffusion model for end-to-end autonomous driving},
  author={Liao, Bencheng and Chen, Shaoyu and Yin, Haoran and Jiang, Bo and Wang, Cheng and Yan, Sixu and Zhang, Xinbang and Li, Xiangyu and Zhang, Ying and Zhang, Qian and others},
  booktitle={Proceedings of the Computer Vision and Pattern Recognition Conference},
  pages={12037--12047},
  year={2025}
}

@inproceedings{xing2025goalflow,
  title={Goalflow: Goal-driven flow matching for multimodal trajectories generation in end-to-end autonomous driving},
  author={Xing, Zebin and Zhang, Xingyu and Hu, Yang and Jiang, Bo and He, Tong and Zhang, Qian and Long, Xiaoxiao and Yin, Wei},
  booktitle={Proceedings of the Computer Vision and Pattern Recognition Conference},
  pages={1602--1611},
  year={2025}
}

@article{zhou2026autovla,
  title={Autovla: A vision-language-action model for end-to-end autonomous driving with adaptive reasoning and reinforcement fine-tuning},
  author={Zhou, Zewei and Cai, Tianhui and Zhao, Seth and Zhang, Yun and Huang, Zhiyu and Zhou, Bolei and Ma, Jiaqi},
  journal={Advances in Neural Information Processing Systems},
  volume={38},
  pages={27920--27956},
  year={2026}
}

@inproceedings{renz2025simlingo,
  title={Simlingo: Vision-only closed-loop autonomous driving with language-action alignment},
  author={Renz, Katrin and Chen, Long and Arani, Elahe and Sinavski, Oleg},
  booktitle={Proceedings of the Computer Vision and Pattern Recognition Conference},
  pages={11993--12003},
  year={2025}
}

@article{de2019causal,
  title={Causal confusion in imitation learning},
  author={De Haan, Pim and Jayaraman, Dinesh and Levine, Sergey},
  journal={Advances in neural information processing systems},
  volume={32},
  year={2019}
}

@article{le2022survey,
  title={A survey on imitation learning techniques for end-to-end autonomous vehicles},
  author={Le Mero, Luc and Yi, Dewei and Dianati, Mehrdad and Mouzakitis, Alexandros},
  journal={IEEE Transactions on Intelligent Transportation Systems},
  volume={23},
  number={9},
  pages={14128--14147},
  year={2022},
  publisher={IEEE}
}

@article{kiran2021deep,
  title={Deep reinforcement learning for autonomous driving: A survey},
  author={Kiran, B Ravi and Sobh, Ibrahim and Talpaert, Victor and Mannion, Patrick and Al Sallab, Ahmad A and Yogamani, Senthil and P{\'e}rez, Patrick},
  journal={IEEE transactions on intelligent transportation systems},
  volume={23},
  number={6},
  pages={4909--4926},
  year={2021},
  publisher={IEEE}
}

@inproceedings{zhang2022rethinking,
  title={Rethinking closed-loop training for autonomous driving},
  author={Zhang, Chris and Guo, Runsheng and Zeng, Wenyuan and Xiong, Yuwen and Dai, Binbin and Hu, Rui and Ren, Mengye and Urtasun, Raquel},
  booktitle={European Conference on Computer Vision},
  pages={264--282},
  year={2022},
  organization={Springer}
}

@article{gao2026rad,
  title={Rad: Training an end-to-end driving policy via large-scale 3dgs-based reinforcement learning},
  author={Gao, Hao and Chen, Shaoyu and Jiang, Bo and Liao, Bencheng and Shi, Yiang and Guo, Xiaoyang and Pu, Yuechuan and Li, Xiangyu and Liu, Wenyu and Zhang, Qian and others},
  journal={Advances in Neural Information Processing Systems},
  volume={38},
  pages={32551--32576},
  year={2026}
}

@article{ni2025recondreamer,
  title={Recondreamer-rl: Enhancing reinforcement learning via diffusion-based scene reconstruction},
  author={Ni, Chaojun and Zhao, Guosheng and Wang, Xiaofeng and Zhu, Zheng and Qin, Wenkang and Chen, Xinze and Jia, Guanghong and Huang, Guan and Mei, Wenjun},
  journal={arXiv preprint arXiv:2508.08170},
  year={2025}
}

@inproceedings{chen2019model,
  title={Model-free deep reinforcement learning for urban autonomous driving},
  author={Chen, Jianyu and Yuan, Bodi and Tomizuka, Masayoshi},
  booktitle={2019 IEEE intelligent transportation systems conference (ITSC)},
  pages={2765--2771},
  year={2019},
  organization={IEEE}
}

@article{chekroun2023gri,
  title={Gri: General reinforced imitation and its application to vision-based autonomous driving},
  author={Chekroun, Raphael and Toromanoff, Marin and Hornauer, Sascha and Moutarde, Fabien},
  journal={Robotics},
  volume={12},
  number={5},
  pages={127},
  year={2023},
  publisher={MDPI}
}

@article{zhang2016query,
  title={Query-efficient imitation learning for end-to-end autonomous driving},
  author={Zhang, Jiakai and Cho, Kyunghyun},
  journal={arXiv preprint arXiv:1605.06450},
  year={2016}
}

@inproceedings{prakash2020exploring,
  title={Exploring data aggregation in policy learning for vision-based urban autonomous driving},
  author={Prakash, Aditya and Behl, Aseem and Ohn-Bar, Eshed and Chitta, Kashyap and Geiger, Andreas},
  booktitle={Proceedings of the IEEE/CVF Conference on Computer Vision and Pattern Recognition},
  pages={11763--11773},
  year={2020}
}

@article{popov2024mitigating,
  title={Mitigating covariate shift in imitation learning for autonomous vehicles using latent space generative world models},
  author={Popov, Alexander and Degirmenci, Alperen and Wehr, David and Hegde, Shashank and Oldja, Ryan and Kamenev, Alexey and Douillard, Bertrand and Nist{\'e}r, David and Muller, Urs and Bhargava, Ruchi and others},
  journal={arXiv preprint arXiv:2409.16663},
  year={2024}
}

@article{liu2024curse,
  title={Curse of rarity for autonomous vehicles},
  author={Liu, Henry X and Feng, Shuo},
  journal={nature communications},
  volume={15},
  number={1},
  pages={4808},
  year={2024},
  publisher={Nature Publishing Group UK London}
}

@article{wymann2000torcs,
  title={Torcs, the open racing car simulator},
  author={Wymann, Bernhard and Espi{\'e}, Eric and Guionneau, Christophe and Dimitrakakis, Christos and Coulom, R{\'e}mi and Sumner, Andrew},
  journal={Software available at http://torcs. sourceforge. net},
  volume={4},
  number={6},
  pages={2},
  year={2000}
}

@article{li2022metadrive,
  title={Metadrive: Composing diverse driving scenarios for generalizable reinforcement learning},
  author={Li, Quanyi and Peng, Zhenghao and Feng, Lan and Zhang, Qihang and Xue, Zhenghai and Zhou, Bolei},
  journal={IEEE transactions on pattern analysis and machine intelligence},
  volume={45},
  number={3},
  pages={3461--3475},
  year={2022},
  publisher={IEEE}
}

@inproceedings{yang2023unisim,
  title={Unisim: A neural closed-loop sensor simulator},
  author={Yang, Ze and Chen, Yun and Wang, Jingkang and Manivasagam, Sivabalan and Ma, Wei-Chiu and Yang, Anqi Joyce and Urtasun, Raquel},
  booktitle={Proceedings of the IEEE/CVF Conference on Computer Vision and Pattern Recognition},
  pages={1389--1399},
  year={2023}
}

@inproceedings{tonderski2024neurad,
  title={Neurad: Neural rendering for autonomous driving},
  author={Tonderski, Adam and Lindstr{\"o}m, Carl and Hess, Georg and Ljungbergh, William and Svensson, Lennart and Petersson, Christoffer},
  booktitle={Proceedings of the IEEE/CVF Conference on Computer Vision and Pattern Recognition},
  pages={14895--14904},
  year={2024}
}

@software{alpasim_2025,
  author       = {
    NVIDIA and
    Yulong Cao and
    Riccardo de Lutio and
    Sanja Fidler and
    Guillermo Garcia Cobo and
    Zan Gojcic and
    Maximilian Igl and
    Boris Ivanovic and
    Peter Karkus and
    Janick Martinez Esturo and
    Marco Pavone and
    Aaron Smith and
    Ellie Tanimura and
    Michal Tyszkiewicz and
    Michael Watson and
    Qi Wu and
    Le Zhang
  },
  title        = {AlpaSim: A Modular, Lightweight, and Data-Driven Research Simulator for Autonomous Driving},
  year         = {2025},
  month        = {October},
  url          = {https://github.com/NVlabs/alpasim},
}

@article{garcia2025road,
  title={RoaD: Rollouts as Demonstrations for Closed-Loop Supervised Fine-Tuning of Autonomous Driving Policies},
  author={Garcia-Cobo, Guillermo and Igl, Maximilian and Karkus, Peter and Zhang, Zhejun and Watson, Michael and Chen, Yuxiao and Ivanovic, Boris and Pavone, Marco},
  journal={arXiv preprint arXiv:2512.01993},
  year={2025}
}

@inproceedings{zhang2025closed,
  title={Closed-loop supervised fine-tuning of tokenized traffic models},
  author={Zhang, Zhejun and Karkus, Peter and Igl, Maximilian and Ding, Wenhao and Chen, Yuxiao and Ivanovic, Boris and Pavone, Marco},
  booktitle={Proceedings of the Computer Vision and Pattern Recognition Conference},
  pages={5422--5432},
  year={2025}
}

@article{zhao2026bridgesim,
  title={BridgeSim: Unveiling the OL-CL Gap in End-to-End Autonomous Driving},
  author={Zhao, Seth Z and Wang, Luobin and Ruan, Hongwei and Bao, Yuxin and Chen, Yilan and Leng, Ziyang and Ravichandran, Abhijit and He, Honglin and Zhou, Zewei and Han, Xu and others},
  journal={arXiv preprint arXiv:2604.10856},
  year={2026}
}

@article{hu2023gaia,
  title={Gaia-1: A generative world model for autonomous driving},
  author={Hu, Anthony and Russell, Lloyd and Yeo, Hudson and Murez, Zak and Fedoseev, George and Kendall, Alex and Shotton, Jamie and Corrado, Gianluca},
  journal={arXiv preprint arXiv:2309.17080},
  year={2023}
}

@article{russell2025gaia,
  title={Gaia-2: A controllable multi-view generative world model for autonomous driving},
  author={Russell, Lloyd and Hu, Anthony and Bertoni, Lorenzo and Fedoseev, George and Shotton, Jamie and Arani, Elahe and Corrado, Gianluca},
  journal={arXiv preprint arXiv:2503.20523},
  year={2025}
}

@inproceedings{yang2025drivearena,
  title={Drivearena: A closed-loop generative simulation platform for autonomous driving},
  author={Yang, Xuemeng and Wen, Licheng and Wei, Tiantian and Ma, Yukai and Mei, Jianbiao and Li, Xin and Lei, Wenjie and Fu, Daocheng and Cai, Pinlong and Dou, Min and others},
  booktitle={Proceedings of the IEEE/CVF International Conference on Computer Vision},
  pages={26933--26943},
  year={2025}
}

@inproceedings{yan2025drivingsphere,
  title={Drivingsphere: Building a high-fidelity 4d world for closed-loop simulation},
  author={Yan, Tianyi and Wu, Dongming and Han, Wencheng and Jiang, Junpeng and Zhou, Xia and Zhan, Kun and Xu, Cheng-zhong and Shen, Jianbing},
  booktitle={Proceedings of the Computer Vision and Pattern Recognition Conference},
  pages={27531--27541},
  year={2025}
}

@article{chang2025spacer,
  title={SPACeR: Self-Play Anchoring with Centralized Reference Models},
  author={Chang, Wei-Jer and Rangesh, Akshay and Joseph, Kevin and Strong, Matthew and Tomizuka, Masayoshi and Hu, Yihan and Zhan, Wei},
  journal={arXiv preprint arXiv:2510.18060},
  year={2025}
}

@article{distelzweig2026beyond,
  title={Beyond Self-Play and Scale: A Behavior Benchmark for Generalization in Autonomous Driving},
  author={Distelzweig, Aron and Janjo{\v{s}}, Faris and Look, Andreas and Rothenh{\"a}usler, Anna and Jost, Daniel and Scheel, Oliver and Rajan, Raghu and Cornelisse, Daphne and Vinitsky, Eugene and Boedecker, Joschka},
  journal={arXiv preprint arXiv:2605.10034},
  year={2026}
}

@article{qiu2026heterogeneous,
  title={Heterogeneous Self-Play for Realistic Highway Traffic Simulation},
  author={Qiu, Jinkai and Saviolo, Alessandro and Wang, Chaojie and Wang, Mingke and Huang, Xiaoyu},
  journal={arXiv preprint arXiv:2604.16406},
  year={2026}
}

@article{wang2025alpamayo,
  title={Alpamayo-r1: Bridging reasoning and action prediction for generalizable autonomous driving in the long tail},
  author={Wang, Yan and Luo, Wenjie and Bai, Junjie and Cao, Yulong and Che, Tong and Chen, Ke and Chen, Yuxiao and Diamond, Jenna and Ding, Yifan and Ding, Wenhao and others},
  journal={arXiv preprint arXiv:2511.00088},
  year={2025}
}

@book{foley1996computer,
  title={Computer graphics: principles and practice},
  author={Foley, James D},
  volume={12110},
  year={1996},
  publisher={Addison-Wesley Professional}
}

@book{glassner1989introduction,
  title={An introduction to ray tracing},
  author={Glassner, Andrew S},
  year={1989},
  publisher={Morgan Kaufmann}
}

@book{sutton1998reinforcement,
  title={Reinforcement learning: An introduction},
  author={Sutton, Richard S and Barto, Andrew G and others},
  volume={1},
  number={1},
  year={1998},
  publisher={MIT press Cambridge}
}

@inproceedings{saxena2020driving,
  title={Driving in dense traffic with model-free reinforcement learning},
  author={Saxena, Dhruv Mauria and Bae, Sangjae and Nakhaei, Alireza and Fujimura, Kikuo and Likhachev, Maxim},
  booktitle={2020 IEEE International Conference on Robotics and Automation (ICRA)},
  pages={5385--5392},
  year={2020},
  organization={IEEE}
}

@article{harmel2023scaling,
  title={Scaling is all you need: Autonomous driving with jax-accelerated reinforcement learning},
  author={Harmel, Moritz and Paras, Anubhav and Pasternak, Andreas and Roy, Nicholas and Linscott, Gary},
  journal={arXiv preprint arXiv:2312.15122},
  year={2023}
}

@article{jaeger2025carl,
  title={Carl: Learning scalable planning policies with simple rewards},
  author={Jaeger, Bernhard and Dauner, Daniel and Bei{\ss}wenger, Jens and Gerstenecker, Simon and Chitta, Kashyap and Geiger, Andreas},
  journal={arXiv preprint arXiv:2504.17838},
  year={2025}
}

@article{schulman2017proximal,
  title={Proximal policy optimization algorithms},
  author={Schulman, John and Wolski, Filip and Dhariwal, Prafulla and Radford, Alec and Klimov, Oleg},
  journal={arXiv preprint arXiv:1707.06347},
  year={2017}
}

@article{sutton1999policy,
  title={Policy gradient methods for reinforcement learning with function approximation},
  author={Sutton, Richard S and McAllester, David and Singh, Satinder and Mansour, Yishay},
  journal={Advances in neural information processing systems},
  volume={12},
  year={1999}
}

@inproceedings{schulman2015trust,
  title={Trust region policy optimization},
  author={Schulman, John and Levine, Sergey and Abbeel, Pieter and Jordan, Michael and Moritz, Philipp},
  booktitle={International conference on machine learning},
  pages={1889--1897},
  year={2015},
  organization={PMLR}
}

@article{mnih2015human,
  title={Human-level control through deep reinforcement learning},
  author={Mnih, Volodymyr and Kavukcuoglu, Koray and Silver, David and Rusu, Andrei A and Veness, Joel and Bellemare, Marc G and Graves, Alex and Riedmiller, Martin and Fidjeland, Andreas K and Ostrovski, Georg and others},
  journal={nature},
  volume={518},
  number={7540},
  pages={529--533},
  year={2015},
  publisher={Nature Publishing Group}
}

@inproceedings{rowectrlsim,
  title={CtRL-Sim: Reactive and Controllable Driving Agents with Offline Reinforcement Learning},
  author={Rowe, Luke and Girgis, Roger and Gosselin, Anthony and Carrez, Bruno and Golemo, Florian and Heide, Felix and Paull, Liam and Pal, Christopher},
  booktitle={8th Annual Conference on Robot Learning},
  year={2024}
}

@inproceedings{nguyen2026lead,
  title={Lead: Minimizing learner-expert asymmetry in end-to-end driving},
  author={Nguyen, Long and Fauth, Micha and Jaeger, Bernhard and Dauner, Daniel and Igl, Maximilian and Geiger, Andreas and Chitta, Kashyap},
  booktitle={Proceedings of the IEEE/CVF Conference on Computer Vision and Pattern Recognition},
  pages={39775--39785},
  year={2026}
}

@inproceedings{xiao2021pandaset,
  title={Pandaset: Advanced sensor suite dataset for autonomous driving},
  author={Xiao, Pengchuan and Shao, Zhenlei and Hao, Steven and Zhang, Zishuo and Chai, Xiaolin and Jiao, Judy and Li, Zesong and Wu, Jian and Sun, Kai and Jiang, Kun and others},
  booktitle={2021 IEEE international intelligent transportation systems conference (ITSC)},
  pages={3095--3101},
  year={2021},
  organization={IEEE}
}

@inproceedings{geiger2012kitti,
  title={Are we ready for autonomous driving? the kitti vision benchmark suite},
  author={Geiger, Andreas and Lenz, Philip and Urtasun, Raquel},
  booktitle={2012 IEEE conference on computer vision and pattern recognition},
  pages={3354--3361},
  year={2012},
  organization={IEEE}
}

@inproceedings{rowe2025scenario,
  title={Scenario dreamer: Vectorized latent diffusion for generating driving simulation environments},
  author={Rowe, Luke and Girgis, Roger and Gosselin, Anthony and Paull, Liam and Pal, Christopher and Heide, Felix},
  booktitle={Proceedings of the IEEE/CVF Conference on Computer Vision and Pattern Recognition},
  pages={17207--17218},
  year={2025}
}

@inproceedings{isola2017image,
  title={Image-to-image translation with conditional adversarial networks},
  author={Isola, Phillip and Zhu, Jun-Yan and Zhou, Tinghui and Efros, Alexei A},
  booktitle={Proceedings of the IEEE conference on computer vision and pattern recognition},
  pages={1125--1134},
  year={2017}
}

@inproceedings{bousmalis2018using,
  title={Using simulation and domain adaptation to improve efficiency of deep robotic grasping},
  author={Bousmalis, Konstantinos and Irpan, Alex and Wohlhart, Paul and Bai, Yunfei and Kelcey, Matthew and Kalakrishnan, Mrinal and Downs, Laura and Ibarz, Julian and Pastor, Peter and Konolige, Kurt and others},
  booktitle={2018 IEEE international conference on robotics and automation (ICRA)},
  pages={4243--4250},
  year={2018},
  organization={IEEE}
}

@inproceedings{chitta2024sledge,
  title={Sledge: Synthesizing driving environments with generative models and rule-based traffic},
  author={Chitta, Kashyap and Dauner, Daniel and Geiger, Andreas},
  booktitle={European Conference on Computer Vision},
  pages={57--74},
  year={2024},
  organization={Springer}
}

@article{coelho2023rlad,
  title={RLAD: Reinforcement learning from pixels for autonomous driving in urban environments},
  author={Coelho, Daniel and Oliveira, Miguel and Santos, Vitor},
  journal={IEEE Transactions on Automation Science and Engineering},
  volume={21},
  number={4},
  pages={7427--7435},
  year={2023},
  publisher={IEEE}
}

@inproceedings{osinski2020simulation,
  title={Simulation-based reinforcement learning for real-world autonomous driving},
  author={Osi{\'n}ski, B{\l}a{\.z}ej and Jakubowski, Adam and Zi{\k{e}}cina, Pawe{\l} and Mi{\l}o{\'s}, Piotr and Galias, Christopher and Homoceanu, Silviu and Michalewski, Henryk},
  booktitle={2020 IEEE international conference on robotics and automation (ICRA)},
  pages={6411--6418},
  year={2020},
  organization={IEEE}
}

@inproceedings{toromanoff2020end,
  title={End-to-end model-free reinforcement learning for urban driving using implicit affordances},
  author={Toromanoff, Marin and Wirbel, Emilie and Moutarde, Fabien},
  booktitle={Proceedings of the IEEE/CVF conference on computer vision and pattern recognition},
  pages={7153--7162},
  year={2020}
}

@article{dauner2024navsim,
  title={Navsim: Data-driven non-reactive autonomous vehicle simulation and benchmarking},
  author={Dauner, Daniel and Hallgarten, Marcel and Li, Tianyu and Weng, Xinshuo and Huang, Zhiyu and Yang, Zetong and Li, Hongyang and Gilitschenski, Igor and Ivanovic, Boris and Pavone, Marco and others},
  journal={Advances in Neural Information Processing Systems},
  volume={37},
  pages={28706--28719},
  year={2024}
}

\clearpage
\appendix

\begin{figure}[p]
    \centering
    \includegraphics[width=\textwidth]{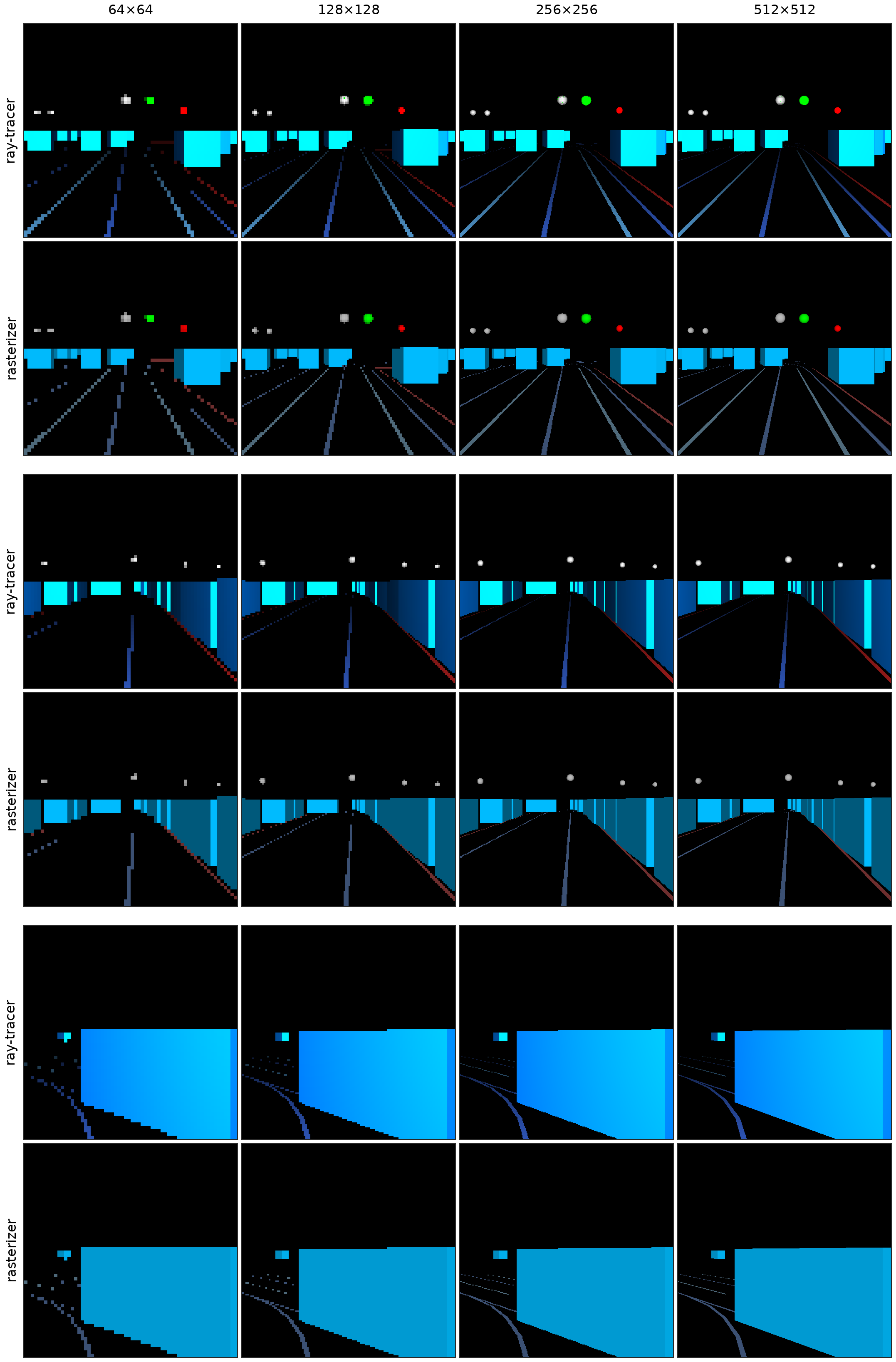}
    \caption{\textbf{Ray-traced vs.\ rasterized Gigapixel renderings.} We render the front-view of three scenes with both the Gigapixel ray-tracer and rasterizer across a range of resolutions. The rasterizer achieves higher throughput, but at the cost of simpler shading, as shown in the bottom two rows.}
    \label{fig:your-label}
\end{figure}

\section{Gigapixel Simulator}

\paragraph{High-Throughput Batched Rendering.}
The Gigapixel simulator extends PufferDrive 2.0 \cite{pufferdrive2025github}, which is built on PufferLib 3.0 \cite{suarez2024pufferlib}. PufferLib 3.0 targets high-throughput RL training on \textit{vectorized} observations, but does not natively support high-throughput rendering for training policies on pixel observations. Gigapixel closes this gap by integrating the GPU-accelerated Madrona batched perspective renderer \cite{rosenzweig2024high}, which supports both rasterized and ray-traced rendering. The Madrona renderer exposes an \emph{entity-component-system} (ECS) abstraction: a scene is a set of \emph{entities}, each described by a few typed \emph{components} (e.g., position, orientation, color), and all rendering logic is expressed as data-parallel \emph{systems} that read and update those components. Crucially, the same systems run across many independent \emph{worlds} in lockstep on a single GPU, so thousands of environments render together in a single batched launch, which is the key to its high rendering throughput.

The driving scene maps onto this abstraction directly. We instantiate one Madrona world per ego agent, each rendering that agent's ego-centric view. Within a world, every renderable element (surrounding vehicles, static obstacles, road segments, and traffic lights) shares a single entity archetype carrying only $\mathrm{SE}(3)$ transform and appearance components; what distinguishes a vehicle from a road segment or a traffic light is the primitive identifier stored in its geometry component. The ego camera and the scene's directional light are themselves entities with their own small component sets. Because we utilize a deliberately simple abstract scene representation, a small set of geometric primitives suffices to realize the scene: vehicles and static objects use cuboids, road segments use planes, and traffic lights use spheres. Shading is minimal, using a single directional light per-agent with per-instance colour (set to blue for all agents and static objects in our experiments).

Rendering uses either a rasterized or a ray-traced backend. The rasterizer is faster, exploiting the simplicity of the primitives, while the ray tracer trades throughput for higher visual fidelity. Once the policy is paired with a compute-intensive architecture such as DrivoR \cite{kirby2026driving}, end-to-end inference and optimization dominate the per-step cost, so rendering is not the practical bottleneck; we therefore use the ray-traced backend as the default unless otherwise stated (see Figure~\ref{fig:training_sps} for a throughput analysis). The full path from simulator state to policy input stays on the GPU: rendered images are exposed directly as a batched GPU tensor to the learning framework, incurring no per-step CPU transfer. Image resolution and camera configuration (including the number of cameras) are configurable, trading visual fidelity against the per-step rendering budget available during self-play. Finally, the renderer scales to distributed training by assigning one GPU per rank. Because rendering is local to each rank and requires no inter-rank communication, throughput scales near-linearly with the number of GPUs.

\paragraph{Observations.}
Each controlled agent receives an ego-centric observation expressed in its own reference frame. The observation combines an ego-state vector, a high-level navigation command, and either rendered pixel observations (for pixel-based policies), or concatenated vectorized features (for vectorized policies).

The \emph{ego state} comprises the signed longitudinal speed, longitudinal and lateral acceleration, the vehicle's length and width, and a binary collision flag. Under the jerk dynamics model it additionally includes the current steering angle, and the longitudinal and lateral accelerations. All continuous quantities are normalized to $[-1,1]$.

We summarize navigation intent with a discrete high-level command: \textsc{left}, \textsc{straight}, \textsc{right}, or \textsc{unknown}. This is the goal representation adopted by end-to-end AV benchmarks such as NAVSIM \cite{cao2025pseudo} and HUGSIM \cite{zhou2025hugsim}, in contrast to conditioning on a dense route or goal position. We derive the command geometrically from the agent's logged trajectory: we snap the agent's current position to the nearest point on its recorded path, advance a fixed look-ahead distance ($20\,\mathrm{m}$) along the path, and transform the resulting target into the egocentric frame. Its lateral offset determines the command, with offsets beyond $\pm 2\,\mathrm{m}$ mapping to \textsc{left} or \textsc{right} and smaller offsets to \textsc{straight}; \textsc{unknown} is emitted when no valid trajectory or snap point exists. The command is available both in self-play and at deployment, and does not expose the dense expert path to the policy.

The vectorized features are: \emph{surrounding agents} (the nearest $K$ within $100\,\mathrm{m}$), each contributing relative position, dimensions, relative heading as $(\cos,\sin)$, and relative signed speed; \emph{road geometry}, encoded as polyline segments retrieved from a spatial hash around the agent, each carrying relative position, length, orientation as $(\cos,\sin)$, and a categorical type (e.g., lane, road edge, crosswalk); and \emph{traffic controls}, giving the relative position and one-hot state (red/yellow/green/unknown) of the nearest signals. When reward conditioning is enabled, the per-agent reward-weight vector is appended to the ego state, allowing a single policy to enact different driving styles.

\paragraph{Action Space and Dynamics.}
The simulator supports two control regimes, depending on whether the policy emits low-level actions or a future trajectory.

\textit{Action-based policies.}
For policies that act directly, we adopt the same action space and dynamics as Gigaflow~\cite{cusumano2025robust}. Each agent is controlled through a discrete set of twelve actions, formed as the Cartesian product of longitudinal jerk values $\{-15, -4, 0, 4\}\,\mathrm{m/s^3}$ and lateral jerk values $\{-4, 0, 4\}\,\mathrm{m/s^3}$, which drive a jerk-actuated kinematic bicycle model. At each step the selected jerks are integrated into longitudinal and lateral accelerations and then into velocity and steering, with accelerations clipped to g-force limits, speed clipped to the speed limit, and the steering angle and its rate bounded to keep trajectories kinematically feasible. The wheelbase is set to $60\%$ of the vehicle length. We refer the reader to Gigaflow~\cite{cusumano2025robust} for the full dynamics derivation.

\textit{Trajectory-based policies.}
Vision-based end-to-end planners such as DrivoR do not emit jerks directly; instead they predict a 4s future trajectory expressed as a sequence of future $\mathrm{SE}(2)$ waypoints $(\Delta x, \Delta y, \Delta\theta)$ in the ego-centric rear-axle frame, replanned at $2\,\mathrm{Hz}$. To execute a planned trajectory in Gigapixel, we adapt the batched LQR controller from NAVSIM~\cite{cao2025pseudo} and run it in a receding-horizon fashion. The predicted waypoints are interpolated to the simulation step ($\Delta t = 0.1\,\mathrm{s}$) to form a dense reference of poses, velocities, and curvatures, which a pair of decoupled LQR controllers tracks: a longitudinal controller maps velocity error to a target acceleration, and a lateral controller maps lateral error, heading error, and steering angle to a steering rate, using a one-second lookahead. The resulting $(\text{acceleration}, \text{steering rate})$ commands are integrated through a kinematic bicycle model, and the propagated pose and velocity are written back to the simulator. For more details on the LQR controller, we refer readers to NAVSIM \cite{cao2025pseudo}.

\paragraph{Reward Functions and Reward Conditioning.} Following Gigaflow~\cite{cusumano2025robust}, the per-step reward is a weighted sum of interpretable terms covering safety, rule compliance, progress, and comfort. The safety terms penalize vehicle-to-vehicle collisions (scaled by impact speed) and leaving the drivable area. The rule-compliance terms penalize red-light violations, driving against the lane direction, exceeding the speed limit, and reversing. Progress is encouraged by a lane-aligned velocity reward and a terminal bonus for reaching the goal, with a small per-step time penalty to discourage stalling. Comfort is promoted by penalizing harsh acceleration and jerk, and a lane-centering term rewards staying near the lane centerline when aligned with it. The functional form of each term is given in Table~\ref{tab:reward}.

To train a single policy that spans a range of driving styles, we randomize the reward coefficients per agent at reset and expose the sampled vector to the policy as an additional observation (\emph{reward conditioning}). Each of the twelve coefficients is drawn independently and uniformly from the ranges in Table~\ref{tab:reward}, normalized to $[-1, 1]$, and appended to the ego state. At inference time, the conditioning vector is fixed to a single nominal ``careful-driver'' setting (rightmost column of Table~\ref{tab:reward}), selecting safe, rule-abiding behavior from the learned family of policies.

\begin{table*}[t]
\centering
\setlength{\tabcolsep}{5pt}
\renewcommand{\arraystretch}{1.4}
\resizebox{\textwidth}{!}{%
\begin{tabular}{@{}l l l c@{}}
\toprule
\textbf{Reward term} & \textbf{Definition} & \textbf{Training distribution} & \textbf{Inference} \\
\midrule
Collision &
  $-(\alpha_{\mathrm{col}} + 0.1\,|v|)\,\mathbbm{1}_{\mathrm{col}}$ &
  $\alpha_{\mathrm{col}} \sim \mathcal{U}(0,\,3)$ & $3.0$ \\
Off-road &
  $-\alpha_{\mathrm{off}}\,\mathbbm{1}_{\mathrm{off\text{-}road}}$ &
  $\alpha_{\mathrm{off}} \sim \mathcal{U}(0,\,3)$ & $3.0$ \\
Comfort &
  $-\alpha_{\mathrm{cmf}}\big(\mathbbm{1}_{|a_{\mathrm{lon}}|>3} + \mathbbm{1}_{|a_{\mathrm{lat}}|>3} + \mathbbm{1}_{|\dot a_{\mathrm{lon}}|>5\,\vee\,|\dot a_{\mathrm{lat}}|>5}\big)$ &
  $\alpha_{\mathrm{cmf}} \sim \mathcal{U}(0,\,0.05)$ & $0.05$ \\
Lane alignment &
  $\alpha_{\mathrm{aln}}\,\Delta t\big(\min(\cos\theta_f,0) + \min(v\cos\theta_f,0) + 0.0025\,(1-\tfrac{2\theta_f}{\pi})\big)$ &
  $\alpha_{\mathrm{aln}} \sim \mathcal{U}(2.5\!\times\!10^{-4},\,2.5\!\times\!10^{-2})$ & $0.025$ \\
Lane centering &
  $-\alpha_{\mathrm{ctr}}\,\Delta t\big(\mathbbm{1}_{\cos\theta_f>0.5}\,|x_f| - 0.05\,e^{\,0.5-|x_f|}\big)$ &
  $\alpha_{\mathrm{ctr}} \sim \mathcal{U}(2.5\!\times\!10^{-4},\,7.5\!\times\!10^{-3})$ & $3.8\!\times\!10^{-3}$ \\
Velocity (progress) &
  $\alpha_{\mathrm{vel}}\,\Delta t\,\max(\cos\theta_f,\,0)\,\mathbbm{1}_{|v|>2.5}$ &
  $\alpha_{\mathrm{vel}} \sim \mathcal{U}(0,\,5\!\times\!10^{-3})$ & $2.5\!\times\!10^{-3}$ \\
Traffic-light &
  $-\alpha_{\mathrm{tl}}\,\mathbbm{1}_{\mathrm{red\text{-}light\ violation}}$ &
  $\alpha_{\mathrm{tl}} \sim \mathcal{U}(0,\,1)$ & $1.0$ \\
Time penalty &
  $-\alpha_{\mathrm{t}}\,\Delta t\,\mathbbm{1}_{|v|>0\,\vee\,|a|>0}$ &
  $\alpha_{\mathrm{t}} \sim \mathcal{U}(0,\,5\!\times\!10^{-5})$ & $2.5\!\times\!10^{-5}$ \\
Reverse &
  $-\alpha_{\mathrm{rev}}\,\Delta t\,\mathbbm{1}_{v<0}$ &
  $\alpha_{\mathrm{rev}} \sim \mathcal{U}(2.5\!\times\!10^{-4},\,7.5\!\times\!10^{-3})$ & $5\!\times\!10^{-3}$ \\
Overspeed &
    $-\alpha_{\mathrm{ovr}}\,\mathbbm{1}_{|v|>v_{\mathrm{lim}}}$ &
    $\alpha_{\mathrm{ovr}} \sim \mathcal{U}(0,\,1)$ & $0.0$ \\
  Goal &
    $\mathbbm{1}_{\lVert x - g\rVert < \delta_{\mathrm{goal}}}$ &
    $\delta_{\mathrm{goal}} \sim \mathcal{U}(2,\,12)$ & $5.0$ \\
  Goal-speed tol. &
    $v_{\mathrm{goal}} = v^{\mathrm{human}}_{\max} + \Delta v_{\mathrm{tol}}$ &
    $\Delta v_{\mathrm{tol}} \sim \mathcal{U}(1,\,5)$ & $5.0$ \\
  \bottomrule
  \end{tabular}%
  }
  \caption{Reward terms, their functional form, and the per-agent randomization used for reward conditioning. Coefficients are positive magnitudes with the sign carried in the definition. $\theta_f$ is
  the heading misalignment with the lane, $x_f$ the lateral offset from the lane center, $v$ the speed, $a_{(\cdot)}$ and $\dot a_{(\cdot)}$ the accelerations and jerks, and $\Delta t$ the simulation
  step. $v^{\mathrm{human}}_{\max}$ is the maximum speed along the agent's human logged trajectory; the goal speed $v_{\mathrm{goal}}$ and the speed limit $v_{\mathrm{lim}}$ are both set to
  $v^{\mathrm{human}}_{\max} + \Delta v_{\mathrm{tol}}$. The rightmost column gives the fixed conditioning used at inference.}
\label{tab:reward}
\vspace{-5mm}
\end{table*}

\paragraph{Other Details.}
The policy controls all vehicles by default, while pedestrians and cyclists follow their logged trajectories. Each episode runs for up to $\!20\,\mathrm{s}$ and terminates at the time limit or once all controlled agents have reached their goal. The goal position is set to the final xy coordinate in the human logs. Scenarios are extracted from nuPlan scenarios; each scenario provides road geometry (lanes, edges, crosswalks, stop signs), traffic-signal state sequences, and recorded trajectories for up to several hundred agents, from which agents' initial states and high-level commands are derived.

On a collision, off-road departure, or red-light violation, the offending agent is halted in place (velocity zeroed) but continues to receive observations. When an agent reaches its goal, it respawns at its initial position in a parallel environment with no surrounding agents. Halting on infraction introduces more static obstacles during training, as rule-violating agents become static obstacles from the perspective of other agents. To make scenes more adversarial and less reactive, a configurable fraction of active agents can be demoted to \emph{log replay}, following their recorded trajectories non-reactively rather than acting under the policy. 

\section{Training Details}

\subsection{Vectorized Teacher Training}
\label{sec:teacher-training}
We train the teacher policy via self-play with decentralized PPO~\cite{schulman2017proximal}. Reward weights are randomized per agent and supplied to the policy as conditioning inputs. We use the high-level discrete navigation command (\textsc{left}/\textsc{straight}/\textsc{right}/\textsc{unknown}) as the goal representation to reduce information asymmetry between the vectorized teacher and pixel-based student. Training takes 24 hours on 8 NVIDIA H200s. The remaining PPO and environment hyperparameters are listed in Table~\ref{tab:ppo-hparams}.

\paragraph{Policy architecture.}
The teacher is a compact ($\sim\!2.7$M-parameter) permutation-invariant network inspired by the Gigaflow architecture~\cite{cusumano2025robust}. The flat observation is split back into its semantic groups: the ego state and the variable-length sets of surrounding agents, road-graph segments, and traffic light states. Each group is embedded by its own two-layer MLP encoder with layer normalization, and the three object sets are aggregated by element-wise max-pooling, yielding a fixed-size, permutation-invariant summary of the scene. The four group embeddings are concatenated and passed through a shared trunk of three GELU-activated, layer-normalized linear blocks, from which linear actor and value heads branch off. The actor produces the factorized (longitudinal, lateral) jerk action. The bulk of the parameters reside in this shared trunk.

\paragraph{Advantage prioritization.}
Driving experience is dominated by routine straight-line driving, while the maneuvers that matter most for driving competence (turns, yields, and near-collision recoveries) are comparatively rare. To focus optimization on these informative transitions, minibatch segments are sampled with probability proportional to the magnitude of their estimated advantage raised to a power $\alpha$, rather than uniformly, upsampling high-magnitude advantage events. Advantage prioritization can be seen as a soft variant of advantage filtering proposed in Gigaflow \cite{cusumano2025robust}.

\begin{table}[t]
\centering
\small
\setlength{\tabcolsep}{8pt}
\renewcommand{\arraystretch}{1.15}
\begin{tabular}{@{}ll@{}}
\toprule
\textbf{Hyperparameter} & \textbf{Value} \\
\midrule
Algorithm                & Decentralized PPO \\
Optimizer                & Adam ($\beta_1{=}0.9$, $\beta_2{=}0.999$, $\epsilon{=}10^{-8}$) \\
Learning rate            & $5\times10^{-4}$ (constant) \\
Discount $\gamma$        & $0.999$ \\
GAE $\lambda$            & $0.95$ \\
PPO clip $\epsilon$      & $0.2$ \\
Value clip               & $0.2$ \\
Value loss coef.         & $0.5$ \\
Entropy coef.            & $0.005$ \\
Max gradient norm        & $1.0$ \\
Batch size               & $524{,}288$ \\
Minibatch size           & $32{,}768$ \\
Segment length              & $32$ \\
Update epochs            & $1$ \\
Advantage-priority $\alpha$ & $0.85$ \\
Importance weight $\beta_0$ & $0.85$ \\
Priority clamp           & $5.0$ \\
\midrule
Agents per map           & $512$ \\
Log-replay fraction      & $0.10$ \\
Episode length           & $67$ steps ($\Delta t{=}0.3\,$s, ${\approx}20\,$s) \\
Training maps (nuPlan)   & $335{,}245$ \\
Total environment steps  & $2.5\times10^{10}$ \\
\bottomrule
\end{tabular}
\vspace{3mm}
\caption{Key hyperparameters for vectorized teacher training. Reward
randomization/conditioning ranges are listed separately in
Table~\ref{tab:reward}.}
\label{tab:ppo-hparams}
\end{table}

\subsection{Pixel-Based Student Training}
\label{sec:student-training}

\begin{table}[t]
\centering
\small
\setlength{\tabcolsep}{8pt}
\renewcommand{\arraystretch}{1.15}
\begin{tabular}{@{}ll@{}}
\toprule
\textbf{Hyperparameter} & \textbf{Value} \\
\midrule
\multicolumn{2}{@{}l}{\emph{Objective}} \\
Training mode            & Self-play DAgger (trajectory + action) \\
Planning loss            & Winner-takes-all (DrivoR) / $L_1$ (DrivoR-Reg) \\
Trajectory loss coef.    & $1.0$ \\
Action (KL) loss coef.   & $1.0$ \\
Cmd loss weights (L/F/R/U) & $5 / 1 / 5 / 1$ \\
Teacher rollout horizon $H$ & $4\,\mathrm{s}$ \\
Teacher mix $\beta$      & $1\!\to\!0$ linear over $125\mathrm{M}$, then $0$ \\
\midrule
\multicolumn{2}{@{}l}{\emph{Architecture}} \\
Vision backbone          & DINOv2 ViT-S/14 (reg4), LoRA \\
Register tokens          & $16$ \\
Decoder depth / width    & $4$ / $1024$ \\
Trajectory proposals     & $64$ (DrivoR) / $1$ (DrivoR-Reg) \\
Waypoints (horizon)      & $8$ (2\,Hz, $4\,\mathrm{s}$) \\
\midrule
\multicolumn{2}{@{}l}{\emph{Optimization}} \\
Optimizer                & AdamW (wd $0.01$) \\
Learning rate            & $2\times10^{-4}$, step $\times0.1$ at $100\mathrm{M}$ \\
Batch / minibatch size   & $8{,}192$ / $256$ \\
Max gradient norm        & $1.0$ \\
DAgger replay capacity   & $25{,}000$ \\
\midrule
\multicolumn{2}{@{}l}{\emph{Environment \& rendering}} \\
Execution (SDC)          & NAVSIM LQR, receding horizon control \\
Execution (non-SDC)      & Student action head, native jerk dynamics \\
Cameras                  & $4$ (front, front-left, front-right, rear) \\
Resolution               & $210{\times}126 \rightarrow 434{\times}252$ (final $25\mathrm{M}$) \\
Episode length           & $201$ steps ($\Delta t{=}0.1\,$s, $\approx20\,$s) \\
Training maps (nuPlan)   & $335{,}245$ \\
Total steps              & $125\mathrm{M} + 25\mathrm{M}$ \\
\bottomrule
\end{tabular}
\vspace{3mm}
\caption{Key hyperparameters for pixel-based student distillation. Reward
randomization/conditioning ranges follow Table~\ref{tab:reward} and matched to the teacher during training.}
\label{tab:student-hparams}
\vspace{-5mm}
\end{table}

We distill the vectorized teacher (Sec.~\ref{sec:teacher-training}) into a pixel-based student via self-play DAgger in Gigapixel: the student observes only rendered images (plus a minimal ego state and high-level navigation command), while the teacher, retaining its privileged vectorized observations, provides on-policy supervision from states the student visits. Training takes 36 hours on 8 H200 GPUs, with training configurations listed in Table~\ref{tab:student-hparams}. We instantiate the student as a DrivoR model \cite{kirby2026driving}. We train two variants: the full multimodal DrivoR planner, supervised with a winner-takes-all loss over its trajectory proposals, and a single-mode regression variant (DrivoR-Reg), supervised with an $L_1$ loss to the teacher trajectory. Relative to the original DrivoR, we make two modifications for compatibility with self-play training: (i)~we augment the student's ego state with vehicle length and width and the reward-conditioning vector so that it can control any agent in self-play, and (ii)~we add a parallel \emph{action head}, driven by a separate action query in the decoder, that predicts the discrete jerk action and steps the student when it controls actors other than the self-driving car (SDC).

\paragraph{Trajectory execution and supervision.}
Teacher trajectory targets $\tau$ are obtained by rolling out the teacher policy $\pi_{\text{teacher}}$ in self-play mode for a horizon of $H=4\,\mathrm{s}$ at $10\,\mathrm{Hz}$ in a forked copy of the simulator. To execute a predicted trajectory in closed loop we use the LQR controller integrated in Gigapixel, but apply it only to the SDC: the controller's kinematic-bicycle parameters are calibrated to nuPlan's self-driving car geometry, so tracking buses or trucks would introduce large tracking errors. All other agents are instead stepped through Gigapixel's native jerk dynamics using the student's parallel action head, avoiding any vehicle-specific re-tuning of the controller. We match the student's reward conditioning vector $\mathbf{c}$ to the teacher, so that the student can be trained to imitate the same persona as the rolled out teacher.

\paragraph{Mixed-policy rollouts.}
Following standard DAgger practice~\cite{ross2011reduction}, we use a mixed-policy rollout schedule: at each step, the executed action is drawn from the teacher with probability $\beta$ and from the student with probability $1-\beta$. We anneal $\beta$ linearly from $1$ to $0$ over the first $125\mathrm{M}$ steps and hold $\beta=0$ thereafter, which limits chaotic self-play dynamics while the student is still converging and gradually shifts the visited-state distribution onto the student's own rollouts.

\paragraph{Data curation.}
Because much of self-play is uninformative straight driving, and because imitating poor-quality trajectories is harmful, we curate the supervision in two ways. First, we reweight the trajectory loss by the high-level navigation command, upweighting the rarer left- and right-turn commands by $5\times$ relative to go-straight to counter the heavy class imbalance toward straight driving. Second, we filter the teacher rollouts that enter the replay buffer, discarding: trajectories whose $H$-second window contains any collision or off-road infraction; trajectories from agents that have already committed an infraction earlier in their current episode; and trajectories truncated before all waypoints are valid (e.g.\ by a goal respawn or episode end), which would otherwise create supervision asymmetry across waypoints.

\paragraph{Rendering and resolution schedule.}
Observations are produced by the Gigapixel ray-traced renderer, which provides front, front-left, front-right, and rear camera views. We train in two phases: the first $125\mathrm{M}$ steps at $210\times126$ resolution, then a final $25\mathrm{M}$ steps at $434\times252$.

\paragraph{Scoring head.}
The scoring-based DrivoR variant ranks trajectory proposals with a learned scorer whose targets are produced by the nuPlan closed-loop simulator. Because the scoring head requires the nuPlan simulator for computing PDMS targets, we do not train the scoring head inside Gigapixel; instead, we first train all components except for the scoring head in Gigapixel, and then train only the scoring head in open-loop on the \texttt{navtrain} split using the DrivoR codebase, keeping all other modules frozen. We refer the reader to DrivoR \cite{kirby2026driving} for the full scorer training details.

\subsection{Sim-to-real Adaptation Training}
\label{sec:sim2real}
The pixel student of Sec.~\ref{sec:student-training} is trained entirely on images from the synthetic Gigapixel renderer. To deploy it on real camera data, we add a final adaptation stage, transferring the policy from rendered observations to real images while preserving the behavior learned in self-play. The Gigapixel-trained student supplies the supervision targets, and we fine-tune the perception stack on real frames under the distillation objective detailed below. Following the DrivoR training configurations, we train the NAVSIM-v2 model for fewer epochs than the HUGSIM model and without the extra \texttt{navval} data: for NAVSIM-v2 we train for $10$ epochs on \texttt{navtrain} ($85\mathrm{k}$ samples), and for HUGSIM we train for $20$ epochs on \texttt{navtrain}\,+\,\texttt{navval} ($103\mathrm{k}$ samples). Both the teacher and the student operate at $434\times252$ resolution. We optimize with AdamW at a learning rate of $2\times10^{-4}$ and a global batch size of $64$ ($16$ per GPU across $4$ GPUs), and set the perceptual feature loss coefficient to  $\lambda=1$. Training takes place on $4$ A100 GPUs in roughly $12$ hours.

\paragraph{Mode-aligned distillation objective.}
For the multimodal DrivoR planner, the teacher and student each output a set of $P=64$ trajectory proposals, $\{\tau_{k}\}_{k=1}^{P}$ and $\{\hat{\tau}_{k}\}_{k=1}^{P}$, where each proposal $\tau_{k}\in\mathbb{R}^{H\times3}$ is a sequence of $H$ future $\mathrm{SE}(2)$ waypoints. During distillation, we supervise the student in a \emph{mode-aligned} fashion: the student's $k$-th proposal is regressed directly toward the teacher's $k$-th proposal, rather than under a winner-takes-all assignment. Whereas DrivoR is normally trained against a single ground-truth target through a winner-takes-all loss $\min_{k}\lVert\hat{\tau}_{k}-\tau^{*}\rVert$, distillation instead applies dense, per-mode supervision against the teacher's full proposal set:
\begin{equation}
\mathcal{L}_{\mathrm{plan}}
= \frac{1}{P}\sum_{k=1}^{P}
\big\lVert \hat{\tau}_{k} - \tau_{k} \big\rVert_{1}.
\end{equation}
The full distillation loss adds an $L_2$ feature-matching term on the decoder register tokens and the scoring head loss:
\begin{equation}
\mathcal{L}
= \lambda\,
   \big\lVert E^{\text{real}}_{\text{per}} - E^{\text{sim}}_{\text{per}} \big\rVert_{2}^{2}
+ \mathcal{L}_{\mathrm{plan}} + \mathcal{L}_{\mathrm{score}},
\end{equation}
where $E^{\text{sim}}_{\text{per}},E^{\text{real}}_{\text{per}}$ are the teacher and student register tokens, and $\mathcal{L}_{\mathrm{score}}$ is the scoring head loss (see DrivoR~\cite{kirby2026driving} for details). For the regression-based DrivoR-Reg variant, no scoring loss is applied and $P=1$, which reduces $\mathcal{L}_{\mathrm{plan}}$ to a single-mode $L_1$ regression toward the teacher trajectory, matching the planning loss used during self-play DAgger training in Gigapixel. 

\section{Benchmark Details}
\label{sec:benchmark-details}
We evaluate in three settings: our own Gigapixel simulator and two real-world driving benchmarks, HUGSIM~\cite{zhou2025hugsim} and NAVSIM-v2~\cite{cao2025pseudo}. Together these probe closed-loop driving on abstract rendered observations (Gigapixel), photorealistic closed-loop driving on reconstructed real scenes (HUGSIM), and pseudo-closed-loop planning on real logs (NAVSIM-v2).

\paragraph{Gigapixel.}
In Gigapixel we evaluate policies on a held-out set of $1{,}000$ scenarios in closed loop, with all surrounding actors replaying their logged trajectories. This setting isolates closed-loop driving performance on Gigapixel's abstract pixel observations, independent of any sim-to-real adaptation. We report the \emph{Gigapixel Driving Score}, $\max\!\big(0,\ \text{Completion Rate} - \text{Collision Rate} - \text{Off-road Rate}\big)$.

\paragraph{HUGSIM.}
HUGSIM~\cite{zhou2025hugsim} is a closed-loop benchmark that reconstructs real-world logs in a 3D Gaussian-splatting simulator, enabling pixel-level closed-loop evaluation on real scenes. It aggregates four source datasets (nuScenes~\cite{caesar2020nuscenes}, Waymo~\cite{sun2020scalability}, PandaSet~\cite{xiao2021pandaset}, and KITTI~\cite{geiger2012kitti}) and reports performance across four difficulty tiers (Easy, Medium, Hard, Extreme) that progressively increase the adversarial behavior of surrounding actors. The official metric is the HUGSIM Driving Score (HD-Score), which multiplies route-completion by a time-averaged, PDMS-style closed-loop score built from safety and comfort subscores (Sec.~\ref{sec:metrics-appendix}). Following DrivoR~\cite{kirby2026driving}, we evaluate on $345$ scenarios.

\paragraph{NAVSIM-v2.}
NAVSIM-v2~\cite{cao2025pseudo} is a pseudo-closed-loop benchmark: planned trajectories are executed non-reactively against rule-based traffic. Its metric is the Extended Predictive Driver Model Score (EPDMS), which multiplicatively combines a Stage 1 score, computed on the real logged poses, with a Stage 2 score, computed on synthetically perturbed poses that probe robustness to off-distribution recovery states (Sec.~\ref{sec:metrics-appendix}). We evaluate on the official leaderboard split, \texttt{navhard}.

\subsection{Driving-Score Metrics}
\label{sec:metrics-appendix}
We provide high-level descriptions of the HUGSIM and NAVSIM-v2 driving scores and refer the reader to the respective papers for the full subscore definitions, weights, and formulas.

\paragraph{HUGSIM HD-Score.}
HUGSIM scores driving with a PDMS-style metric inspired by NAVSIM~\cite{dauner2024navsim}. At each simulation step, the per-step score multiplies two safety subscores, no-collision and drivable-area compliance, by a weighted average of two additional subscores, time-to-collision and comfort. Because the safety terms enter multiplicatively, a collision or off-road event at a step drives that step's score to zero. The final HD-Score averages the per-step score over the episode and multiplies it by a global route-completion score $\text{RC} \in [0,1]$, the fraction of the intended route the policy completes. See HUGSIM~\cite{zhou2025hugsim} for the exact subscore definitions and weights.

\paragraph{NAVSIM-v2 EPDMS.}
NAVSIM-v2 extends the NAVSIM PDMS into the Extended PDMS (EPDMS). As in PDMS, the subscores split into multiplicative \emph{penalty} terms that gate the overall score, no-collision (NC), drivable-area compliance (DAC), driving-direction compliance (DDC), and traffic-light compliance (TLC), and a \emph{weighted-average} term over time-to-collision (TTC), ego progress (EP), lane keeping (LK), history comfort (HC), and extended comfort (EC). Any penalty violation zeroes the score, while the weighted-average term grades overall driving quality. EPDMS is computed in two stages whose scores are combined multiplicatively: Stage 1 evaluates the planner on the real logged poses, and Stage 2 evaluates it on synthetically perturbed poses (novel viewpoints offset from the logged trajectory), measuring robustness when recovering from off-distribution states. We refer the reader to NAVSIM-v2~\cite{cao2025pseudo} for the subscore weights and formulas.

\end{document}